\documentclass{article}

    \PassOptionsToPackage{numbers, compress}{natbib}
    \usepackage[final]{neurips_2021}

\usepackage[utf8]{inputenc} % allow utf-8 input
\usepackage[T1]{fontenc}    % use 8-bit T1 fonts
\usepackage{hyperref}       % hyperlinks
\usepackage{url}            % simple URL typesetting
\usepackage{booktabs}       % professional-quality tables
\usepackage{amsfonts}       % blackboard math symbols
\usepackage{nicefrac}       % compact symbols for 1/2, etc.
\usepackage{microtype}      % microtypography
\usepackage{xcolor}         % colors

\usepackage{algorithm}
\usepackage[noend]{algpseudocode}

\newcommand{\Loss}{\mathcal{L}}

\usepackage{graphicx}
\usepackage[]{todonotes}

\usepackage{tabularx}
\usepackage{comment}

\usepackage{subfig}
\usepackage{caption}
\usepackage{booktabs}

\usepackage{amsmath,amssymb}
\usepackage{rotating}

\usepackage{wrapfig}
\usepackage{multirow}
\usepackage{Definitions}

\newcommand{\bA}{{\bar A}}

\newcommand*\samethanks[1][\value{footnote}]{\footnotemark[#1]}

\title{Noether Networks: \\Meta-Learning Useful Conserved Quantities}

\author{%
  Ferran Alet\thanks{Equal contribution. Our code is publicly available at \href{https://lis.csail.mit.edu/noether}{https://lis.csail.mit.edu/noether}.} $^{\,1}$, Dylan Doblar\samethanks{} $^{\,1}$, Allan Zhou$^{2}$, \\\textbf{Joshua Tenenbaum$^{1}$, Kenji Kawaguchi$^{3}$, Chelsea Finn$^2$}\\%
  $^1$MIT, $^2$Stanford University, $^3$National University of Singapore\\
  \texttt{\{alet,ddoblar\}@mit.edu} \\
}

\begin{document}

\maketitle

\begin{abstract}
Progress in machine learning~(ML) stems from a combination of data availability, computational resources, and an appropriate encoding of inductive biases. Useful biases often exploit symmetries in the prediction problem, such as convolutional networks relying on translation equivariance. Automatically discovering these useful symmetries holds the potential to greatly improve the performance of ML systems, but still remains a challenge. In this work, we focus on sequential prediction problems and take inspiration from Noether's theorem to reduce the problem of finding inductive biases to meta-learning useful conserved quantities.  
We propose Noether Networks: a new type of architecture where a meta-learned conservation loss is optimized inside the prediction function. We show, theoretically and experimentally, that Noether Networks improve prediction quality, providing a general framework for discovering inductive biases in sequential problems.

\end{abstract}

\section{Introduction}
The clever use of inductive biases to exploit symmetries has been at the heart of many landmark achievements in machine learning, such as translation invariance in CNN image classification~\citep{krizhevsky2012imagenet}, permutation invariance in Graph Neural Networks~\citep{scarselli2008graph} for drug design~\citep{stokes2020deep}, or roto-translational equivariance in SE3-transformers~\citep{fuchs2020se} for protein structure prediction~\citep{jumper2021highly}. However, for data distributions of interest, there may be exploitable symmetries that are either unknown or difficult to describe with code.
Progress has been made in automatically discovering symmetries for finite groups~\citep{zhou2020meta}, but meta-learning and exploiting general continuous symmetries has presented a major challenge.
In part, this is because symmetries describe the effect of counterfactuals about perturbations to the data, which are not directly observable.

In this work, we propose to exploit symmetries in sequential prediction problems indirectly.
We take inspiration from Noether's theorem~\cite{Noether1918}, which loosely states the following:
\begin{center}
    \textit{For every continuous symmetry property of a dynamical system, \\there is a corresponding quantity whose value is conserved in time.}
\end{center}
For example, consider a system of planets interacting via gravity: this system is translation invariant in all three cardinal directions (i.e. translating the entire system in the x,y, or z axis conserves the laws of motion). Noether's theorem asserts there must be a conserved quantity for each of these symmetries; in this case, linear momentum. 
Similarly, the system has a time-invariance (i.e. the laws of motion are the same today as they will be tomorrow). In this case, the corresponding conserved quantity is the total energy of the system.

Inspired by this equivalence, we propose that approximate conservation laws are a powerful paradigm for meta-learning useful inductive biases in sequential prediction problems. Whereas symmetries are difficult to discover because they are global properties linked to counterfactuals about unobserved perturbations of the data, conserved quantities can be directly observed in the true data. This provides an immediate signal for machine learning algorithms to exploit.

\begin{figure}
    \centering
    \includegraphics[width=\linewidth]{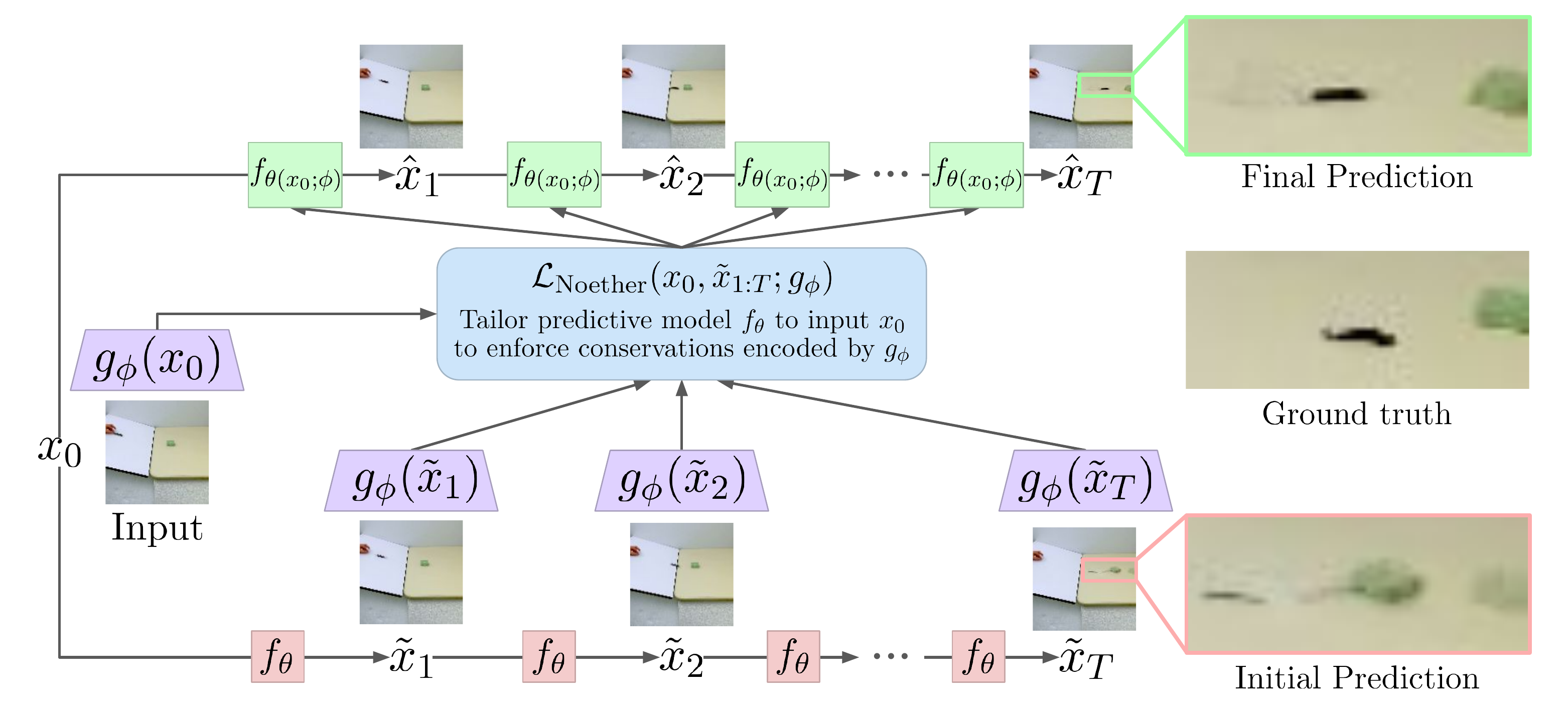}
    \caption{Noether Networks enforce conservation laws, meta-learned by $g_\phi$, in sequential predictions made by $f_\theta$, which is tailored to the input $x_0$ to produce final predictions $\hat{x}_{1:T}$. Imposing these meta-learned inductive biases improves video prediction quality with objects sliding down a ramp.}
    \label{fig:main_diagram}
\end{figure}

Our approach involves meta-learning a parametric conservation loss function which is useful to the prediction task. We leverage the \textit{tailoring} framework~\citep{alet2020tailoring}, which proposes to encode inductive biases by fine-tuning neural networks with hand-designed unsupervised losses inside the prediction function. Whereas traditional auxiliary losses are added to the task loss during training, tailoring losses are optimized inside the prediction function both during training and testing, customizing the model to each individual query. In doing so, we ensure there is no generalization gap for the conservation loss.
We propose to meta-learn the self-supervised loss function and parameterize it in the form of a conservation loss; i.e. $\mathcal{L}(x_{0},\tilde{x}_{1:T}; g_\phi) = \sum_{t=1}^{T}|g_\phi(x_0)-g_\phi(\tilde x_t)|^2$. 
This conservation form encodes a meta-inductive bias (inductive bias over inductive biases) which narrows the search space exponentially (in $T$) and simplifies the parameterization.
Figure~\ref{fig:main_diagram} demonstrates the Noether Network pipeline for a video prediction task.

\iffalse
Our contributions are the following:
\begin{enumerate}
    \item We propose Noether Networks: an architecture class that automatically learns its own inductive biases in the form of meta-learned conservation loss functions.
    \item We show that when the meta-learning of the conservation loss takes the form of program synthesis we can recover known conservation laws from raw physics data.
    \item We highlight the benefits of Noether Networks by improving long-term video prediction by discovering useful conserved quantities from raw pixels.
\end{enumerate}
\fi

The main contribution of this paper is the Noether Network, an architecture class and algorithm that can automatically learn its own inductive biases in the form of meta-learned conservation loss functions. We theoretically characterize the advantages of such conservation laws as effective regularizers that constrain the learning to lower-dimensional output manifolds, lowering the generalization gap. Empirically, we first find that, when the meta-learned conservation loss takes the form of a synthesized program, Noether Networks recover known conservation laws from raw physical data. Second, we find that Noether Networks can learn conservation losses on raw videos, modestly improving the generalization of a video prediction model, especially on longer-term predictions.

%%====================================================================

\section{Theoretical Advantages of Enforcing Conservation Laws} \label{sec:theory}
In this section, we demonstrate principled advantages of enforcing conservation laws of the form $g_\phi(f_{\theta}(x))=g_\phi(x)$ by considering a special case where preimages under $g_\phi$ form affine subspaces.

Let input $x$ and target $y$ be $x,y\in  \RR^d$, and let the Noether embedding %parameterized by $\phi$ 
be $g_\phi: \RR^d \rightarrow \Pcal$ where $\Pcal=\{g_\phi(x): x \in\RR^d\}$. 
% For any $p \in \Pcal$, we define the preimage of $g_\phi$ as $g_\phi^{-1}[\{p\}] = \{x \in  \RR^d : g_\phi(x)=p\}$, where $g_\phi$ is not required to be invertible.
We consider a restricted class of models, parameterized by $\theta\in\Theta$, of the form $f_\theta(x)=x+v_\theta$ for $v_\theta \in \RR^d$  such that for all $x$, the preimage of $g_\phi$ is $g_\phi^{-1}[\{g_\phi(x)\}]=\{x+Az : z \in \RR^m\}$, $A \in \RR^{d\times m}$.
Here, $m \le d$ is the dimensionality of the preimages of $g_\phi$. We denote by
$C$ the smallest upper bound on the loss value as   $\Lcal(f_{\theta}(x),y)\le C$ (for all $x,y$ and $\theta$). Define $\psi(v)= \EE_{x,y}[\Lcal(f(x,v),y)]-\frac{1}{n} \sum_{i=1}^n\Lcal(f(x_i,v),y_i)$, with Lipschitz constant $\zeta$. Therefore, $\zeta \ge 0$  is the smallest real number such that, for all $v$ and $v'$ in $\Vcal$, $|\psi(v) -\psi(v')| \le \zeta  \|v - v'\|_2$, where $\Vcal=\{v_\theta \in \RR^d : \theta \in \Theta\}$.
Finally, we define
$R=\sup_{\theta\in\Theta} \|v_\theta\|_{2}$, and 
the generalization gap by 
\[
  G(\theta)=\EE_{x,y}[\Lcal(f_{\theta}(x),y)]-\frac{1}{n} \sum_{i=1}^n\Lcal(f_\theta(x_i),y_i).
\]
Theorem \ref{thm:2} shows that enforcing conservation laws of $g_\phi(f_{\theta}(x))=g_\phi(x)$  is advantageous when the dimension of the preimages under $g_\phi$ is less than the dimension of $x$; that is, when $m<d$.

\def\kkswitch{1}
\if\kkswitch0

\begin{theorem} \label{thm:1}
(without conservation laws)\ Let $\rho \in \NN^+$. Then, for any $\delta>0$,  with  probability at least $1-\delta$ over an
iid draw of $n$ examples $((x_i
, y_i))_{i=1}^n$, the following holds for all $\theta \in \Theta$:
\begin{align}\label{eq:genbound1}
G(\theta) \le C \sqrt{\frac{d\ln (\sqrt{d})+d\ln (2R(\zeta^{1-1/\rho})\sqrt{n})+\ln(1/\delta)}{2n}}+ \sqrt{\frac{\zeta^{2/\rho}}{n}}.  
\end{align}
\end{theorem}

\begin{theorem} \label{thm:2}
(with conservation laws) Let $\rho \in \NN^+$. If $g_\phi(f_{\theta}(x))=g_\phi(x)$ for any $x \in \RR^d$ and $\theta \in \Theta$, then for any $\delta>0$,  with  probability at least $1-\delta$ over an
iid draw of $n$ examples $((x_i
, y_i))_{i=1}^n$, the following holds for all $\theta \in \Theta$:
\begin{align}\label{eq:genbound2}
G(\theta) \le C \sqrt{\frac{m\ln (\max(\sqrt{m},1))+m\ln (2R(\zeta^{1-1/\rho})\sqrt{n})+\ln(1/\delta)}{2n}}+\one\{m\ge1\} \sqrt{\frac{\zeta^{2/\rho}}{n}}.  
\end{align} 
\end{theorem}

\else

\begin{theorem} \label{thm:2}
Let $\rho \in \NN^+$. Then, for any $\delta>0$,  with  probability at least $1-\delta$ over an
iid draw of $n$ examples $((x_i
, y_i))_{i=1}^n$, the following holds for all $\theta \in \Theta$:
\begin{align}\label{eq:genbound2}
G(\theta) \le C \sqrt{\frac{\xi\ln (\max(\sqrt{\xi},1))+\xi\ln (2R(\zeta^{1-1/\rho})\sqrt{n})+\ln(1/\delta)}{2n}}+\one\{\xi\ge1\} \sqrt{\frac{\zeta^{2/\rho}}{n}}.  
\end{align} 
where $\xi=m$ if $g_\phi(f_{\theta}(x))=g_\phi(x)$ for any $x \in \RR^d$ and $\theta \in \Theta$, and $\xi=d$ otherwise.
\end{theorem}
  
\fi

The proof %of Theorem \ref{thm:2} 
is presented in Appendix \ref{app:proofs}.
In Theorem \ref{thm:2}, when we enforce the conservation laws, $d$ is replaced by $m$, the dimension of the preimage.
We now discuss various cases for the values of $m$: \vspace{-8pt}
\begin{itemize}
\item 
(Case of $m=d$) Let us consider an extreme scenario where the function $g_\phi$ maps all $x \in \RR^d$ to one single point. In this scenario, the dimensionality of preimages under $g_\phi$ is maximized as $m=d$. Accordingly, the  bounds with and without enforcing the conservation laws become the same. %in Theorem \ref{thm:2}. 
Indeed, in this scenario, the conservation laws of $g_\phi(f_{\theta}(x))=g_\phi(x)$ give us no information, because they always hold for all $x \in \RR^d$, even without imposing them.
\item 
(Case of $m=0$) Let us consider another extreme scenario where the function $g_\phi$ is invertible. 
In this scenario, the dimensionality of preimages under $g_\phi$ is zero as $m=0$. Thus, imposing the condition of $g_\phi(f_{\theta}(x))=g_\phi(x)$ makes the bound in Theorem \ref{thm:2} to be very small. 
Indeed, in this scenario, the condition of $g_\phi(f_{\theta}(x))=g_\phi(x)$ implies that $f_{\theta}(x)=x$: i.e., $x$ is not moving, and thus it is easy to generalize. 
\item 
(Case of $0<m<d$) From the above two cases, we can see that the benefit of enforcing conservation laws
 comes from more practical cases in-between these, with $0<m<d$. 
\end{itemize}
In Theorem \ref{thm:2}, the function $g_\phi$ can differ from the true  function  $g_\phi^*$ underlying the system. This is because we analyze a standard generalization gap: i.e., the difference between the expected loss and the training loss. The cost of not using the true $g_\phi^{*}$ is captured in the training loss; i.e., the training loss can be large with the function $g_\phi$ that differs significantly from the true $g_\phi^{*}$. Even in this case, the generalization gap can be small. For example, in the case of $m=0$, the generalization bound is small, whereas the training loss will be large unless $x_{t+1}=x_{t}$. Therefore, our analysis gives us the insight on the trade-off between the training loss and the dimensionality of preimages under $g_\phi$.

%%====================================================================

\section{Noether Networks}\label{sec:algorithm}

\paragraph{Leveraging tailoring to encode inductive biases.}
We perform a prediction-time optimization to encourage outputs to follow conservation laws using the tailoring framework~\citep{alet2020tailoring}.
Tailoring encodes inductive biases in the form of unsupervised losses optimized inside the prediction function.
In doing so, tailoring fine-tunes the model to each query to ensure that it minimizes the unsupervised loss for that query.
For example, we may optimize for energy conservation in a physics prediction problem. 
In meta-tailoring, we train the model to do well on the task loss after the tailoring step has fine-tuned its parameters. 
In contrast to auxiliary losses, which would optimise the conservation loss only for the training points, tailoring allows us to ensure conservation at test time. Since we aim to build in the conservation in the architecture, we want to ensure it is also satisfied for unseen test samples. 
Another advantage of tailoring losses 
is that they are easier to meta-learn. Auxiliary losses are pooled over all examples and training epochs and their effect is only known at validation/test time. We would need to use implicit gradients~\citep{lorraine2020optimizing,raghu2020teaching} to know their eventual effect on the final weights at the end of training. With tailoring, we can directly measure the effect of the meta-learned update on the same sample.

A limitation of tailoring framework is that the tailoring loss must be user-specified.
This is acceptable in domains where the desired inductive bias is both known and easily encoded, but problematic in general --- we address this issue with Noether Networks.
Our approach can be seen as a generalization of tailoring where the unsupervised loss is meta-learned and takes the form of a conservation loss.~\looseness=-1

\paragraph{Noether Networks: discovering \textit{useful} conservation laws.}
We build on the paradigm of conservation laws; i.e. quantities that are conserved over time. This has the following two challenges:
\begin{enumerate}
    \item Real data often has noise, breaking any exact conservation law. Moreover, many conservation laws are only approximately satisfied given only partial information of the system. For example, conservation of energy is not fully satisfied in real dissipative systems and estimating conservation of mass from pixels is bound to be inexact in the case of occlusions.
    \item Optimizing for conservation alone can lead to trivial quantities, such as predicting a constant value $g_\phi(x) = C$ independent of $x$.
\end{enumerate}

We search for \textit{useful} conservation laws, whose (approximate) enforcing brings us closer to the true data manifold for the current sequence. Note that a useful conserved quantity doesn't need to be exactly conserved in the training data to improve prediction performance. We only need its conservation to be informative of the data manifold. This allows us to discover conservation laws even in imperfect, noisy environments. Conversely, tailoring with a trivial conserved quantity cannot improve the final predictions (see App.~\ref{app:trivial} for a detailed explanation). Finally, viewing conserved quantities as useful inductive biases aligns well with their use, since we often care about biases only insofar as they help improve task performance.

The search for such useful conserved quantities can take many forms. In this work, we focus on two: first, in the beginning of section~\ref{subsec:exp_ideal}, we use a combination of program synthesis and gradient descent to generate a large, but finite, number of candidate parametric formulas for physical system prediction. We then try meta-tailoring with each formula on the training set and pick the conservation formula with the best loss. Formulas can be a useful, interpretable description in scientific domains given the appropriate state-space. However, for general descriptions from raw data, we would like to describe the loss with a neural network, as we describe in the next section.

\paragraph{Meta-learning a neural loss function.}
We propose Noether Networks, an architecture class for sequential prediction that consists of a base predictor $f_\theta:\tilde{x}_t\mapsto \tilde{x}_{t+1}$ and a meta-learned \mbox{tailoring} loss, parameterized as the conservation of a neural embedding $g_\phi:\tilde{x}_t\mapsto \mathbb{R}^k$. This embedding takes raw predictions as input (such as images in the case of video prediction).
The conservation form induces a meta-inductive bias over potential learned tailoring losses.
The Noether loss is formulated as\looseness=-1
\begin{equation}
  \mathcal{L}_{\rm Noether}(x_0, \tilde{x}_{1:T}; g_\phi) = \underbrace{\sum_{t=1}^T \left|g_\phi(x_0)-g_\phi\left(\tilde{x}_t\right)\right|^2}_{\rm (a)} \approx \underbrace{\sum_{t=1}^T \left|g_\phi(\tilde{x}_{t-1})-g_\phi\left(\tilde{x}_t\right)\right|^2}_{\rm (b)} \label{eq:noether_zero_loss}
\end{equation}
where $x_0$ is the ground-truth input, the $\tilde{x}_t=f_{\theta}(\tilde{x}_{t-1})$ are the model's predictions, 
% $g_\phi$ is the embedding function parameterized by $\phi$,
and $\tilde{x}_0\triangleq x_0$.
% \begin{equation}
%   \mathcal{L}_{\rm Noether}(x_0, \tilde{x}_{1:T}; \phi) = \sum_{t=1}^T \left|g_\phi(\tilde{x}_{t-1})-g_\phi\left(\tilde{x}_t\right)\right|^2\label{eq:noether_prev_loss}
% \end{equation}
Expressions \ref{eq:noether_zero_loss}(a) and \ref{eq:noether_zero_loss}(b) are equivalent if we fully enforce the conservation law, but they differ if conservation is not fully enforced.
When not fully conserving, \ref{eq:noether_zero_loss}(a) propagates information from ground truth more directly to the prediction, but \ref{eq:noether_zero_loss}(b) may be a useful approximation which better handles imperfectly conserved quantities, where the quantity should be Lipschitz but not exactly constant.
In both cases, if we tailor $\theta$ with a single gradient step; the gradient update takes the form
\begin{equation}
  \theta(x_0; \phi) = \theta - \lambda_{\rm in}\nabla_\theta\mathcal{L}_{\rm Noether}(x_0, \tilde{x}_{1:T}(\theta); g_\phi).
\end{equation}
We  compute final predictions as $\hat{x}_t = f_{\theta(x_0;\phi)}(\hat{x}_{t-1})$.
We can now backpropagate %the task loss
from $\mathcal{L}_{\rm task}(x_{1:T}, \hat{x}_{1:T})$ back to $\phi$, which will be optimized so that the unsupervised adaptation $\theta\mapsto \theta(x_0;\phi)$ helps lower $\mathcal{L}_{\rm task}$.% when the sequence starts at $x_0$. 
The optimization requires second-order gradients to express how $\phi$ affects $\mathcal{L}_{\rm task}$ through $\theta(x_0;\phi)$. This is similar to MAML~\citep{finn2017model}, as well as works on meta-learning loss functions for few-shot learning~\citep{antoniou2019learning} and group distribution shift~\citep{zhang2020adaptive}. Algorithm~\ref{alg:noether-meta-train} provides pseudo-code.

\begin{algorithm}[t]
    \caption{Prediction and training procedures for Noether Networks with neural conservation loss}
    % \caption{Prediction procedure for Noether Networks with neural conservation loss}
    \label{alg:noether-meta-train}
    % \hspace*{\algorithmicindent} \textbf{Input:} base model $f$, its un-tailored parameters $\theta$, Noether embedding $g_\phi$, input state $x_0$,\\
    % \hspace*{\algorithmicindent}\hspace*{\algorithmicindent}\hspace*{\algorithmicindent} prediction horizon $T$, learning rate $\lambda_{\rm in}$. \\
    % \hspace*{\algorithmicindent} \textbf{Output:} Final predicted sequence $\hat{x}_{1:T}$.
    \begin{algorithmic}[1] % The number tells where the line numbering should start
        \Statex \textbf{Given:} predictive model class $f$; embedding model class $g$; prediction horizon $T$; training dist.~$\mathcal{D}_{\rm train}$; batch size $N$; 
        % \hspace*{112pt}%
        % \hspace*{\algorithmicindent}%
        learning rates $\lambda_{\rm in}$, $\lambda_{\rm out}$, $\lambda_{\rm emb}$; task loss $\Loss_{\rm task}$; Noether loss $\Loss_{\rm Noether}$
        % \Statex
        \vspace{6pt}
        \Procedure{PredictSequence}{$x_0;\theta,\phi$}%{$f,\theta,g,\phi,x_0,T,\lambda_{\rm in}$} % \Comment{$x_{1:c+p}^{(n)}$: batch of seqs.}
            % \State $\phi,\theta\gets$ randomly initialized weights \Comment{Initialize conserved embedding and model weights}
            % \While{ not done}
                % \State  Sample batch $x_{0:T}^{(0)},\ldots,x_{0:T}^{(N)}\sim \mathcal{D}_{\rm eval}$; assign
                    % $\tilde{x}^{(n)}_{0}, \hat{x}^{(n)}_{0} \gets x^{(n)}_{0}, x^{(n)}_{0}\;\forall n$
                % \ForAll{ $x_{0:T}$}
                % \For{$0\le n\le N$} %\Comment{$N$ batches}
                    %\State $\theta_\gamma\gets \mathbf{1}$
                    %\State $\theta_\beta\gets \mathbf{0}$
                    % \State $\theta^0 \gets \theta$ \Comment{$\theta^s$ denotes the $s^{\rm th}$ set of tailored weights}
                    % \For{$1\le s\le{\rm num\_inner\_steps}$}
                    % \State $\ell\gets 0$  \Comment{Cumulative inner loss for (batched) sequence}
                    %\State $\hat x_{1:c}^{(n)} \gets x_{1:c}^{(n)}$ \Comment{Notational convenience for unrolling}
                    % \State $\tilde{x}^{(n)}_{0}, \hat{x}^{(n)}_{0} \gets x^{(n)}_{0}, x^{(n)}_{0}$
                    %\For{$1\le t\le T$} \Comment{unroll the time-series prediction}
                        \State $\tilde{x}_{0}, \hat{x}_{0} \gets x_{0}, x_{0}$
                        \State $\tilde x_{t}\gets f_{\theta}(\tilde x_{t-1})\;\forall t\in\{1,\ldots,T\}$
                     \Comment{Initial predictions}
                        % \State $\ell \gets \ell + {\rm MSE}(g_\phi(\hat x_{t}^{(n)}), g_\phi(\hat x_{t-1}^{(n)}))$ \Comment{Conserve embedding between time-steps}
                        % \State $\ell \gets \ell + {\rm MSE}(g_\phi(\hat x_{t}^{(n)}), g_\phi(x_{0}^{(n)}))$ \Comment{Conserve embedding between time-steps}
                    % \EndFor
                    % \State $\ell \gets \sum_{t=1}^T {\rm MSE}(g_\phi(\hat x_{t}^{(n)}), g_\phi(x_{0}^{(n)}))$ \Comment{Cumulative inner loss for sequence}
                    % \State $\theta_{x_0^{(n)}}\gets \theta - \lambda_{\rm in}\nabla_{\theta} \ell$
                    \State $\theta(x_0;\phi)\gets \theta - \lambda_{\rm in}\nabla_{\theta} \mathcal{L}_{\rm Noether}(x_0, \tilde{x}_{1:T}; g_\phi)$ \Comment{Inner step with Noether loss}
                    %\State $\theta_{\gamma}^s \gets \theta_{\gamma}^{s-1} - \lambda_{in}\nabla_{\theta_\gamma^{s-1}} \ell$ \Comment{Different elements do \textit{not} influence each other's updates.}
                    %\State $\theta_{\beta}^s \gets \theta_{\beta}^{s-1} - \lambda_{in}\nabla_{\theta_\beta^{s-1}} \ell$
                % \EndFor
                % \State $\hat{x}^{(n)}_{0} \gets x^{(n)}_{0}$
                % \For{$1\le t\le T$}
                \State $\hat x_{t}\gets f_{\theta({x_0};\phi)}(\hat{x}_{t-1})\;\forall t\in\{1,\ldots,T\}$ \Comment{Final prediction with tailored weights}
                % \EndFor
                %\Comment{Optionally, optimize $\theta$ w.r.t.\ outer loss}
                % \EndFor
                % \State $\phi \gets \phi - \lambda_{\rm emb} \nabla_{\phi} \sum_{n=0}^N\Loss_{\rm task}(\hat x_{1:T}^{(n)}, x_{1:T}^{(n)})$ \Comment{Outer step with task loss}
                % \State $\theta \gets \theta - \lambda_{\rm out} \nabla_{\theta}\sum_{n=0}^N \Loss_{\rm task}(\hat x_{1:T}^{(n)}, x_{1:T}^{(n)})$ \Comment{Meta-tailoring outer step}% (omit if tailoring)}
            % \EndWhile
            \State \textbf{return} $\hat{x}_{1:T}$
        \EndProcedure
        % \Statex
        \vspace{4pt}
        % \Statex \textbf{Input:} base model $f$, Noether embedding $g$, training distribution $\mathcal{D}_{\rm train}$, batch size $N$,
        % \Statex \hspace*{\algorithmicindent} prediction horizon $T$, learning rates $\lambda_{\rm in},\lambda_{\rm out},\lambda_{\rm emb}$.
        % \Statex \textbf{Output:} Final predicted sequence $\hat{x}_{1:T}$.
        \Procedure{Train}{}%{$f,g,\mathcal{D}_{\rm train},N,T,\lambda_{\rm in},\lambda_{\rm out},\lambda_{\rm emb}$} % \Comment{$x_{1:c+p}^{(n)}$: batch of seqs.}
            \State $\phi\gets$ randomly initialized weights \Comment{Initialize weights for Noether embedding $g$}
            \State $\theta\gets$ randomly initialized weights \Comment{Initialize weights for predictive model $f$}
            \While{ not done}
                \State  Sample batch $x_{0:T}^{(0)},\ldots,x_{0:T}^{(N)}\sim \mathcal{D}_{\rm train}$%; assign $\tilde{x}^{(n)}_{0}, \hat{x}^{(n)}_{0} \gets x^{(n)}_{0}, x^{(n)}_{0}\;\forall n$
                % \ForAll{ $x_{0:T}$}
                \For{$0\le n\le N$} %\Comment{$N$ batches}
                    % \State $\hat{x}_{1:T}^{(n)} \gets \textsc{PredictSequence}(f,\theta,g,\phi,x_0^{(n)},T,\lambda_{\rm in})$
                    \State $\hat{x}_{1:T}^{(n)} \gets \textsc{PredictSequence}(x_0^{(n)};\theta,\phi)$
                    % \State $\tilde x_{t}^{(n)}\gets f_{\theta}(\tilde x_{t-1}^{(n)})\;\forall t\in\{1,\ldots,T\}$
                    % \Comment{Initial predictions}
                    % \State $\theta_{x_0^{(n)}}\gets \theta - \lambda_{\rm in}\nabla_{\theta} \mathcal{L}_{\rm Noether}(x_0^{(n)}, \tilde{x}_{1:T}^{(n)}; g_\phi)$ 
                    % \Comment{Inner step with Noether loss}
                    % \State $\hat x_{t}^{(n)}\gets f_{\theta_{x_0^{(n)}}}(\hat{x}_{t-1}^{(n)})\;\forall t\in\{1,\ldots,T\}$
                    % \Comment{Final prediction with tailored weights}
                \EndFor
                %\Comment{Optionally, optimize $\theta$ w.r.t.\ outer loss}
                % \EndFor
                \State $\phi \gets \phi - \lambda_{\rm emb} \nabla_{\phi} \sum_{n=0}^N\Loss_{\rm task}(\hat x_{1:T}^{(n)}, x_{1:T}^{(n)})$ \Comment{Outer step with task loss for embedding}
                \State $\theta \gets \theta - \lambda_{\rm out} \nabla_{\theta}\sum_{n=0}^N \Loss_{\rm task}(\hat x_{1:T}^{(n)}, x_{1:T}^{(n)})$ \Comment{Outer step for predictive model}% (omit if tailoring)}
            \EndWhile
            \State \textbf{return} $\phi,\theta$
        \EndProcedure
    \end{algorithmic}
\end{algorithm}

For deep learning frameworks that allow per-example weights, such as JAX~\citep{jax2018github}, the loop over sequences in Alg.~\ref{alg:noether-meta-train} can be efficiently parallelized. To parallelize it for other frameworks we use the \textsc{CNGrad} algorithm~\citep{alet2020tailoring}, which adapts only the Conditional Normalization~(CN) layers in the inner loop. Similar to BN layers, CN only performs element-wise affine transformations: $y = x\odot\gamma + \beta$, which can be efficiently parallelized in most deep learning frameworks even with per-example $\gamma,\beta$.

Finally, even though we use the two-layer optimization typical of meta-learning, we are still in the classical single-task single-distribution setting. Noether Networks learn to impose their own inductive biases via a learned loss and to leverage it via an adaptation of its parameters $\theta$.
This is useful as we often do not have access to meta-partitions that distinguish data between distributions or tasks.

\section{Experiments}~\label{sec:experiments}
Our experiments are designed to answer the following questions:
\begin{enumerate}
    \item Can Noether Networks recover known conservation laws in scientific data?
    \item Are Noether Networks useful for settings with controlled dynamics?
    \item Can Noether Networks parameterize useful conserved quantities from raw pixel information?
    \item How does the degree of conservation affect performance on the prediction task?
\end{enumerate}
\subsection{Experimental domains}
To answer the above questions, we consider a number of different experimental domains, which we overview in this subsection, before moving on to the evaluation and results.
\paragraph{Spring and pendulum from state coordinates.}\label{subsec:exp_ideal}
\citet{greydanus2019hamiltonian} propose the setting of an ideal spring and ideal pendulum, which will allow us to understand the behavior of Noether Networks for scientific data where we know a useful conserved quantity: the energy. They also provide data from a real pendulum from~\citep{schmidt2009distilling}. In contrast to the ideal setups, here the data is noisy and the Hamiltonian is only approximately conserved, as the system is dissipative.
For the two pendulum, input $x=(p,q)\in\mathbb{R}^2$ contains its angle $q$ and momentum $p$. Given the parameters in~\citep{greydanus2019hamiltonian} the energy is $\mathcal{H} = 3(1-\cos{q}) + p^2$, with $p^2 - 3\cdot\cos{q}$ being a simpler equivalent.
For the spring, input $x=(p,q)\in\mathbb{R}^2$ contains the displacement $q$ and momentum $p$ and the energy is $\mathcal{H} = \frac{1}{2}(q^2 + p^2)$. Thus, $q^2+1.0\cdot p^2$ is a conserved quantity, where the coefficient $1.0$ has appropriate units.\looseness=-1

We build on their vanilla MLP baseline and discover conservation laws that, when used for meta-tailoring, improve predictions. Since the baseline predicts $\frac{d}{dt}(x_t)$ rather than $x_{t+1}$, we apply the loss to a finite-difference approximation, i.e. $\mathcal{L}_{\rm Noether}\left(x_t, x_t+\frac{dx}{dt}\Delta_t\right) = \mathcal{L}_{\rm Noether}\left(x_t, x_t+f_\theta(x_t)\Delta_t\right)$.

\paragraph{Domain specific language for scientific formulas.}\label{subsec:DSL}
To search over Hamiltonians, we program a simple domain specific language~(DSL) for physical formulas. Since formulas in the DSL have physical meaning, each sub-formula carries its own associated physical units and is checked for validity.
This allows us to significantly prune the exponential space, as done in AI-Feynman~\cite{udrescu2020ai}. The vocabulary of the DSL is the following: \verb|Input(i)|: returns the $i$-th input, \verb|Operation|: one of $\{+,-,\cdot,/,\sin,\cos,x^2\}$, \verb|Parameter(u)|: trainable scalar with units $[u]$.

%\iffalse
\begin{table}
    \parbox{.495\linewidth}{
    \centering
    \caption{Recovering $\mathcal{H}$ for the ideal pendulum.}
    \begin{tabular}{ccc}
        \toprule[1.5pt]
        Method & Description & RMSE \\
        \midrule[2pt]
        Vanilla MLP & N/A & $0.0563$\\ 
        Noether Nets & $p^2 - 2.99\cos(q)$ & $0.0423$ \\
        \midrule
        True $\mathcal{H}$ [oracle] & $p^2 -3.00 \cos(q)$ & $0.0422$ \\ 
        \bottomrule[1.5pt]
    \end{tabular}
    % \caption{Recovering $\mathcal{H}$ for the ideal pendulum.}
    \label{tab:ideal_pend}
    }
    \parbox{0.495\linewidth}{
    \centering
    \caption{Recovering $\mathcal{H}$ for the ideal spring.}
    \begin{tabular}{ccc}
        \toprule[1.5pt]
        Method & Description & RMSE\\
        \midrule[2pt]
        Vanilla MLP & N/A & $.0174$ \\ 
        Noether Nets & $q^2 + 1.002\text{ } p^2$& $.0165$ \\
        \midrule
        True $\mathcal{H}$ [oracle] & $q^2 +1.000\text{ }p^2$ & $.0166$ \\ 
        \bottomrule[1.5pt]
    \end{tabular}
    % \caption{Recovering $\mathcal{H}$ for the ideal spring.}
    \label{tab:ideal_spring}
    }
\end{table}

\paragraph{Pixel pendulum with controls.}
In settings such as robotics we may be interested in \textit{action-conditioned} video prediction \citep{oh2015action}: predicting future frames given previous frames and an agent's actions. We recorded videos of episodes in OpenAI Gym's pendulum swing up environment, using a scripted policy. %to control the pendulum. 
Each model receives four history frames and a sequence of $26$ policy actions starting from the current timestep, and must predict $26$ future frames. As this is a visually simple environment  
we limit the training dataset to $5$ episodes of $200$ frames each, holding out 
$195$ episodes for testing.

\paragraph{Video prediction with real-world data.}
To characterize the benefits provided by Noether Networks in a real-world video prediction setting, we evaluate the effect of adding a Noether conservation loss to the stochastic video generation (SVG) model~\cite{denton2018stochastic} on the ramp scenario of the Physics 101 dataset~\citep{wu2016physics}.  The dataset contains videos of various real-world objects sliding down an incline and colliding with a second object.  Each model is conditioned on the first two frames and must predict the subsequent 20 frames.  To increase difficulty, we restrict our data to the 20-degree incline. 

\subsection{Evaluation}\label{sec:evaluation}

\paragraph{Can Noether Networks recover known conservation laws in scientific data?}
%The number of formulas increases exponentially with the number of terms involved, leading to vast search spaces. To save computation, 
We first generate all valid formulas up to depth 7, removing operations between incompatible physical units and formulas equivalent up to commutativity. We obtain 41,460 and 72,372 candidate conserved formulas for the pendulum and spring, respectively, which need \textit{not} have units of energy.% since in general not all conserved quantities are energies.
% {% https://tex.stackexchange.com/questions/313967/how-to-place-a-wrapfigure-on-top-of-the-page
% \parfillskip=0pt
% \parskip=0pt
% \par}
%\iffalse
\begin{figure}
\begin{minipage}[c]{0.4\textwidth}
\caption{Noether Networks can recover the energy of a real pendulum, even though it is not fully conserved. This is because they only look for quantities whose conservation helps improve predictions. Moreover, by only softly encouraging conservation, it better encodes imperfect conservations.}\label{fig:real_pendulum}
\end{minipage}
\begin{minipage}[c]{0.59\textwidth}
{\includegraphics[width=\linewidth]{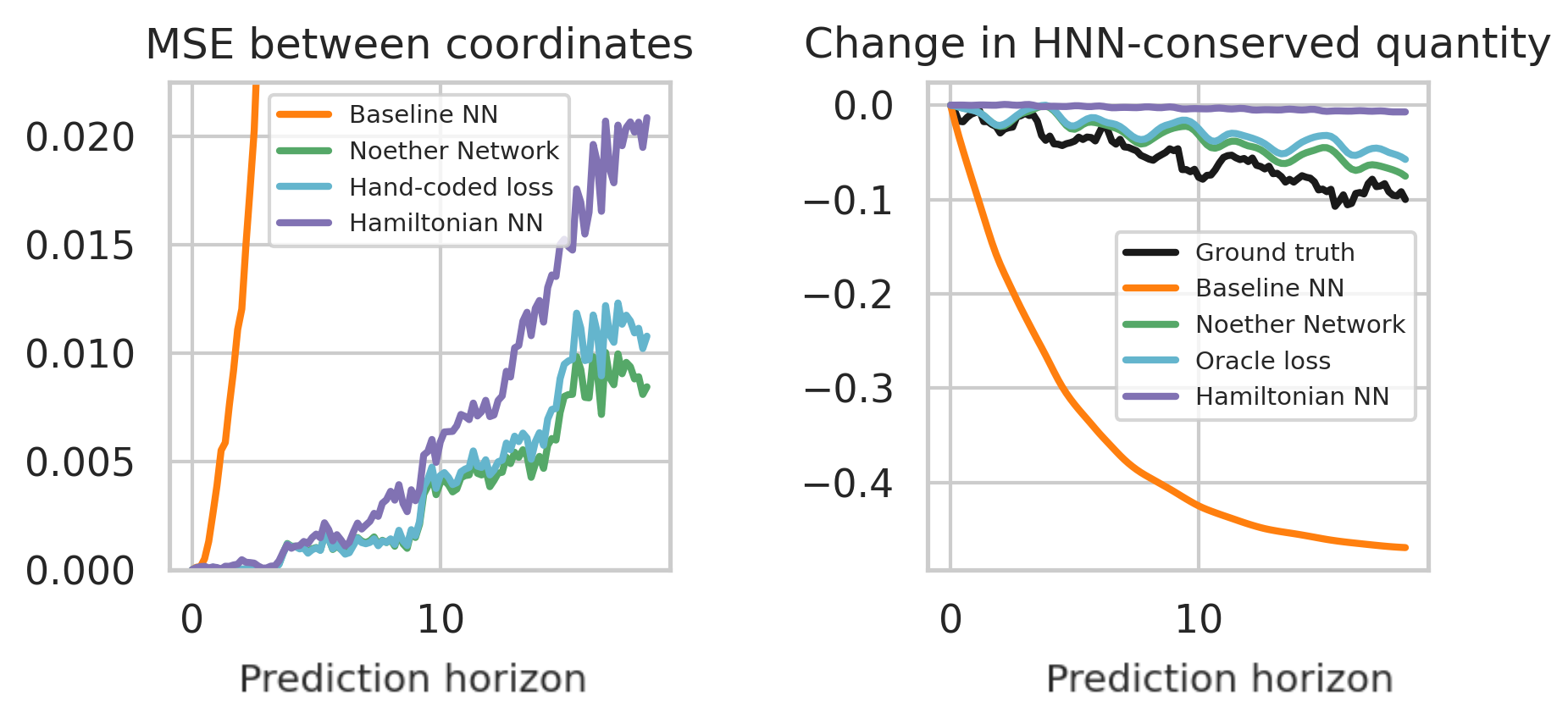}}

    %\vspace{-4pt}
\end{minipage}
\vspace{-5mm}
\end{figure}
%\fi
\iffalse
\begin{wrapfigure}{r}{0.6\linewidth}
 \centering
    % \vspace{-3mm}
    \includegraphics[width=\linewidth]{figures/noether_real_pend_white_background_v2.png}
    \vspace{-5mm}
    \caption{Noether Networks can recover the energy of a real pendulum, even though it is not fully conserved. This is because they only look for quantities whose conservation helps improve predictions. Moreover, by only softly encouraging conservation, it better encodes imperfect conservations.}
    \vspace{-4pt}
    \label{fig:real_pendulum}
\end{wrapfigure}
\fi
% \noindent%
These formulas are purely symbolic, with some trainable parameters still to be defined. We thus optimize their parameters via gradient descent to have low variance within sequences of true data. We then do the same for random sequences; if conservation is two orders of magnitude smaller for the true data, we accept it as an approximate conservation. This measure is similar to others used before~\citep{liu2020ai}, but is not sufficient when the data is noisy, the minimisation is sub-optimal, or there are numerical issues. As a result, we obtain 210 and 219 candidates for approximately conserved quantities for the pendulum and spring, respectively.\looseness=-1
 
Finally, for each potentially conserved quantity, we try it as meta-tailoring loss: starting from the pre-trained vanilla MLP, we fine-tune it for 100 epochs using meta-tailoring, with one inner step and a range of inner learning rates $10^{k}$ for $k\in\{-3,-2.5,\ldots,1\}$. 
We then evaluate the fine-tuned model on long-term predictions, keeping the expression with the best MSE loss in the training data.
We observe that this process correctly singles out equations of the same form as the true Hamiltonian $\mathcal{H}$, with almost the exact parameters. 
Using these as losses reaches equivalent performance with that of the oracle, which uses the true formula (see Tables~\ref{tab:ideal_pend} and~\ref{tab:ideal_spring}).

Finally, we run the same process for a real pendulum~\citep{schmidt2009distilling}, where energy is not conserved. We use largely the same pipeline as for the ideal pendulum, the differences are explained in Appendix~\ref{app:scientific}. Noether Networks discover $p^2-2.39\cos(q)$, close to the (potentially sub-optimal) $\mathcal{H} = p^2-2.4\cos(q)$ described in \citep{greydanus2019hamiltonian}. This Noether Network improves the baseline by more than one order of magnitude, matches the performance of hand-coding the conservation loss, and improves over Hamiltonian Neural Networks, which fully impose conservation of energy. Results can be seen in figure~\ref{fig:real_pendulum}.~\looseness=-1

\paragraph{Are Noether Networks useful in settings with controlled dynamics?}
% \begin{figure}[b]
\begin{figure}
\begin{minipage}[c]{0.49\textwidth}
\caption{%
In controlled pendulum environment, the Noether Network has lower mean squared error (left) and slightly better structural similarity (right). Metrics are computed and plotted by prediction timestep, where timestep $0$ is a given history frame. Note that simply training SVG for more steps does not increase performance.}
    \label{fig:ac-metrics}
\end{minipage}
\begin{minipage}[c]{0.50\textwidth}
    \includegraphics[width=\linewidth]{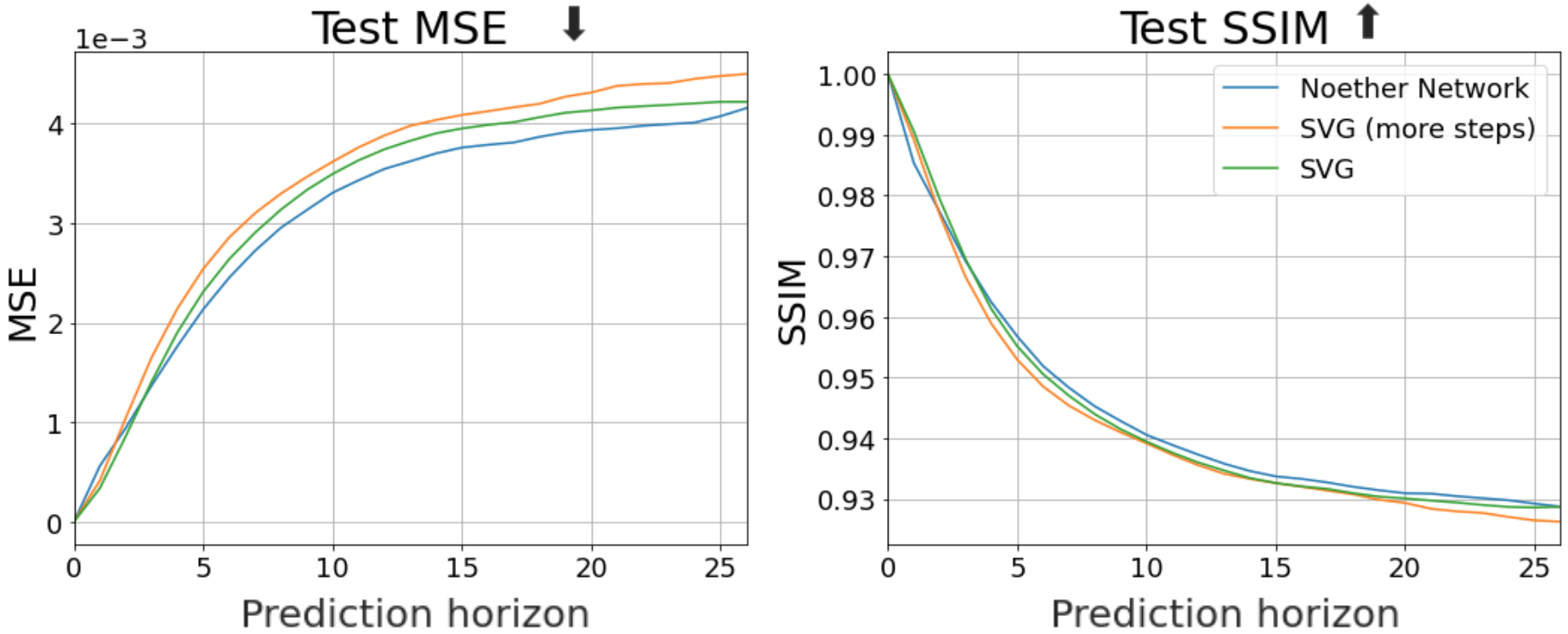}
\end{minipage}
\vspace{-5mm}
\end{figure}
\iffalse
\begin{wrapfigure}{r}{0.6\linewidth}
    \centering
    \vspace{-10pt}
    % TODO(ddoblar): add down/up arrows to plot titles 
    \includegraphics[width=\linewidth]{figures/ac_test_v2.png}
    \caption{Test metrics in the controlled pendulum environment. The Noether Network has lower mean squared error (left) and slightly better structural similarity (right). Metrics are computed and plotted by prediction timestep, where timestep $0$ is a given history frame. Note that simply training SVG for more steps does not increase performance.}
    \label{fig:ac-metrics}
    \vspace{-6pt}
\end{wrapfigure}
\fi
% \end{figure}
For the controlled pendulum experiments, we begin with an SVG model but modify it to (1) remove the prior sampling component since the environment is deterministic and (2) concatenate each action to the corresponding frame encoder output and feed the result to the LSTM. After training the SVG model for 50 epochs and fixing its weights, we run Algorithm \ref{alg:noether-meta-train} for 20 epochs to learn an embedding network for conserving useful quantities. We found that Noether Networks perform better in this setting by directly adapting the latent LSTM activations against the conservation loss, rather than adapting any network parameters. Since Noether Networks take additional training steps starting from a pre-trained SVG model, we also continued training the SVG model for 20 further epochs, which we call ``SVG (more steps).'' Figure~\ref{fig:ac-metrics} shows the results of each method on held out sequences. Noether Networks improve the overall mean squared error (MSE) and structural similarity (SSIM) over the base SVG.~\looseness=-1

\begin{figure}[t]
    \centering
    \includegraphics[width=\textwidth]{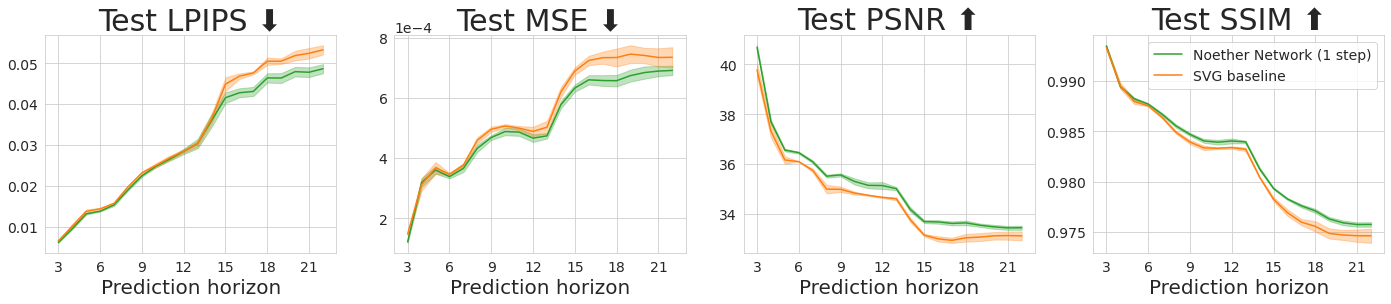}
    \caption{The Noether Network outperforms the baseline by a small margin in all four metrics in real-world video prediction, showing they can meta-learn useful conserved quantities from raw pixel data.~\looseness=-1}
    \label{fig:phys101-metrics}
\vspace{-5mm}    
\end{figure}
\vspace{-3mm}
\paragraph{Can Noether Networks parameterize useful conserved quantities from raw pixel information?}
We train on only 311 training sequences, which makes generalization difficult.  
In this setting, the base SVG model struggles with overfitting --- novel test objects often morph into objects seen at train-time, and the object in motion often morphs with the target object (as shown in figure~\ref{fig:main_diagram}).  Our Noether Network uses the inner loss formulation of Equation \ref{eq:noether_zero_loss}(b), where $g_\phi$ is a two-layer CNN receiving two consecutive frames followed by a fully-connected projection layer producing 64-dimensional embeddings.
We meta-tailor the embedding and the base model for 400 epochs.
As seen in figure~\ref{fig:phys101-metrics}, taking a single inner tailoring step improves performance slightly over the baseline SVG model with respect to all four considered metrics: learned perceptual image patch similarity (LPIPS)~\citep{zhang2018perceptual}, mean squared error (MSE), peak signal-to-noise ratio (PSNR), and structural similarity (SSIM).
As the prediction horizon increases, the Noether Network performs comparatively better, likely because conservations encourage the model to retain aspects of the ground-truth input.
These results provide some evidence that Noether Networks are capable of learning useful inductive biases from raw video data.~\looseness=-1

% \begin{wrapfigure}{l}{0.5\linewidth}
\begin{figure}[t]
    \centering
    \begin{minipage}{0.02\textwidth}
    \begin{turn}{90}
        $\longleftarrow$ Prediction horizon
    \end{turn}
    \end{minipage}%
    \hfill
    \begin{minipage}{0.48\textwidth}
    \centering
    (a) Orange object
    \includegraphics[width=\linewidth]{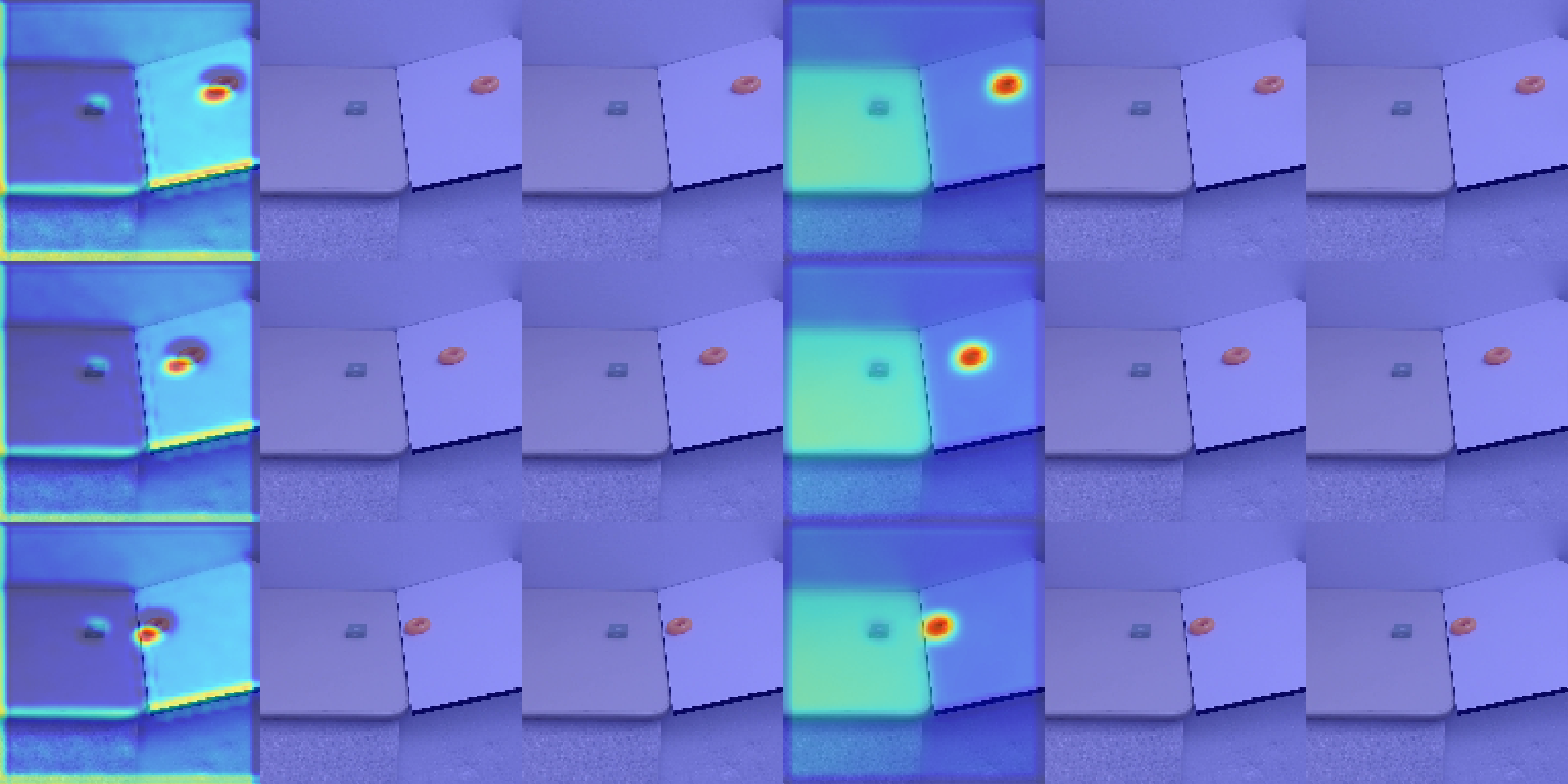}
    Embedding dimension (after PCA) $\longrightarrow$
    \end{minipage}%
    \hfill
    \begin{minipage}{0.48\textwidth}
    \centering
    (b) Blue object
    \includegraphics[width=\linewidth]{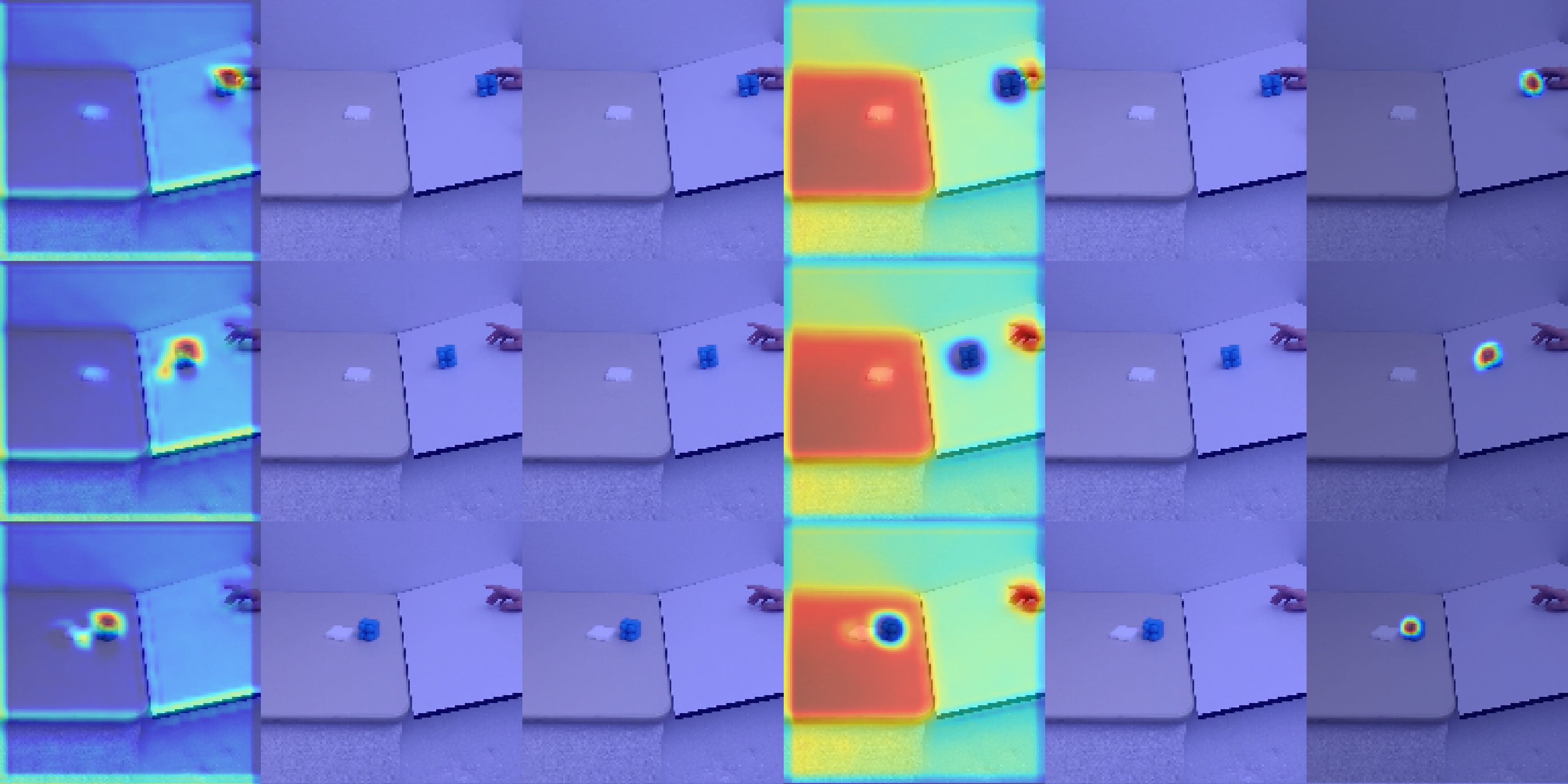}
    Embedding dimension (after PCA) $\longrightarrow$
    \end{minipage}%
    \caption{Grad-CAM localization maps show that the Noether embeddings attend to relevant frame regions for two example sequences.  Heatmaps for the first six dimensions of the embedding (in the PCA basis, sorted by decreasing amount of variance explained) are shown for three time steps.}
    \label{fig:phys101-grad-cam}
\vspace{-5mm}    
\end{figure}
% \end{wrapfigure}

To investigate whether the Noether embeddings learn relevant features, we use Gradient-weighted Class Activation Mapping (Grad-CAM)~\citep{selvaraju2017grad} to compute importance maps for sequences in the test set. % which show frame regions that are ``important'' to each dimension of the embedding.
Since the Noether embedding has 64 dimensions, we perform principal component analysis~(PCA) before Grad-CAM to reduce the dimensionality of the embedding, and to sort dimensions by the percentage of variance they explain in the test set frames.
Interestingly, we find that the first PCA dimension captures 83.6\% of the variance, and the first four dimensions capture 99.9\% of the variance.

Figure~\ref{fig:phys101-grad-cam} shows Grad-CAM localization maps for two example
test-set sequences, (a) depicts an orange torus and (b) shows a blue brick, where warmer colors (red) indicate high importance and cooler colors (blue) indicate low importance. Our interpretations of Grad-CAM localization maps are consistent across examples, see Appendix~\ref{app:grad_cam} for additional ones.
The first dimension, which explains the vast majority of the variance, primarily focuses on the sliding object in both examples.
It also attends to the object on the table, and to the edge of the ramp.
The attention to the objects suggests that the Noether Network learns to conserve quantities related to the objects' pixels and their motion (since it takes{% https://tex.stackexchange.com/questions/313967/how-to-place-a-wrapfigure-on-top-of-the-page
\parfillskip=0pt
\parskip=0pt
\par}
\begin{wrapfigure}{r}{0.33\linewidth}
    \vspace{-4pt}
    \centering
    \includegraphics[width=\linewidth]{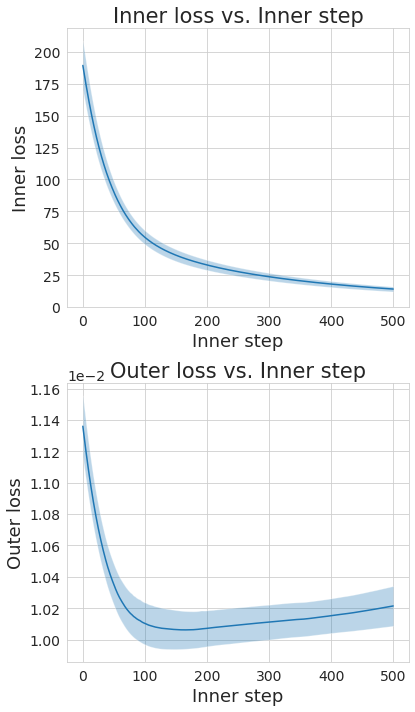}
    \caption{Despite training with only one inner step, as we lower the Noether loss to zero (top), the outer loss remains low (bottom).~\looseness=-1}
    \label{fig:phys101-inner-outer}
    \vspace{-14mm}
\end{wrapfigure}
\noindent%
two consecutive frames).
We hypothesize that the attention to the table edge encodes information about the orientation of the sequence, since half of the training sequences are randomly flipped horizontally.
While some the dimensions focus nearly uniformly on the entire frame, the fourth dimension focuses on the orange sliding object in (a), and the human hand and table in (b).
In all the test set examples, whenever the human hand is in the frame, it is attended to by this dimension.
We hypothesize that it learns to conserve the hand in sequences where it is present, and possibly picks up on similarly-colored pixels as well. Finally, the sixth dimension seems to track blue sliding objects.\looseness=-1

\paragraph{How does the degree of conservation affect performance?}
All of the results presented in this work were achieved by Noether Networks trained and evaluated with a single inner step.
To characterize how the degree to which the conservations are imposed affects video prediction performance, we optimize the inner (Noether) loss for many inner steps and measure the outer (task) loss, as shown in figure~\ref{fig:phys101-inner-outer}.
Here, the inner loss is optimized by Adam for 500 steps, as opposed to the single step of SGD during training (both settings use a learning rate of $10^{-4}$).
During the optimization, the outer loss improves for roughly 150 inner steps, which suggests that the approximate conservations learned by the Noether Network generalize to more exact conservation settings.\looseness=-1

\vspace{-3mm}
\section{Related Work}\label{sec:RW}
\vspace{-2mm}
\paragraph{Unsupervised adaptation and meta-learning loss functions.} Hand-designed unsupervised losses have been used to adapt to distribution shifts with rotation-prediction~\citep{sun2019test} or entropy-minimization~\citep{wang2020fully}, as well as to encode inductive biases~\citep{alet2020tailoring}. Unsupervised losses have also been meta-learned for learning to encode human demonstrations~\citep{yu2018one}, few-shot learning exploiting unsupervised information~\citep{metz2018meta,antoniou2019learning}, and learning to adapt to group distribution shifts~\citep{zhang2020adaptive}. In contrast, our unsupervised loss only takes the single query we care about, thus imposing no additional assumptions on top of standard prediction problems, and takes the form of a conservation law for predicted sequences.\looseness=-1

\vspace{-2mm}
\paragraph{Encoding of symmetries and physics-based inductive biases in neural networks.} We propose to encode symmetries in dynamical systems as meta-learned conserved quantities. This falls under the umbrella of geometric deep learning, which aims to encode symmetries in neural networks;~\citet{bronstein2021geometric} provide an excellent review of the field. Most relevant to the types of problems we focus on is roto-translational equivariance~\citep{thomas2018tensor,weiler20183d,fuchs2020se,satorras2021n}, applications of GNNs in physical settings~\citep{chang2016compositional,battaglia2016interaction,kipf2018neural,alet2019neural,alet2019graph,rossi2020temporal,pfaff2020learning,sanchez2020learning,de2020combining} and the encoding of time-invariance in RNNs~\citep{elman1990finding,jordan1997serial,hochreiter1997long,cho2014learning,tallec2018can}.
Recent works have encoded Hamiltonian and Lagrangian mechanics into neural models~\citep{greydanus2019hamiltonian,lutter2019deep,cranmer2020lagrangian,finzi2020simplifying}, with gains in data-efficiency in physical and robotics systems, including some modeling controlled or dissipative systems~\citep{zhong2019symplectic,desai2021port}. In contrast to these works, we propose to encode biases via prediction-time fine-tuning following the tailoring framework~\citep{alet2020tailoring}, instead of architectural constraints. This allows us to leverage the generality of loss functions to encode a class of inductive biases extending beyond Lagrangian mechanics, and handle raw video data. Outside mechanics,~\citet{suh2020surprising} highlight the difficulty of learning physical inductive biases with deep models from raw data and instead propose to encode mass conservation from pixels with a constrained linear model. Finally, Noether's theorem has been used~\citep{gluch2021noether,tanaka2021noether} to theoretically understand the optimization dynamics during learning; unrelated to our goal of discovering inductive biases for sequential prediction.

\vspace{-2mm}
\paragraph{Discovery of symmetries and conservation laws.} There are several previous works which aim to learn conserved quantities in dynamical systems from data.  The approach of \citet{schmidt2009distilling} focuses on candidate equations that correctly predict dynamic relationships between parts of the system by measuring the agreement between the ratios of the predicted and observed partial (time) derivatives of various components of a system.  A set of analytical expressions is maintained and updated as new candidates achieve sufficient accuracy. More recently,~\citet{liu2020ai} discover conservation laws of Hamiltonian systems by performing local Monte Carlo sampling followed by linear dimensionality estimation to predict the number of conserved quantities from an explained ratio diagram.~\citet{wetzel2020discovering} learn a Siamese Network that learns to differentiate between trajectories. Contrary to both, Noether Networks do not need segmentations into trajectories with different conserved quantities and can deal with raw pixel inputs and outputs. Other approaches to learning symmetries from data include neuroevolution for learning connectivity patterns \citep{stanley2009hypercube}, regularizers for learning data augmentation \citep{benton2020learning}, and meta-learning equivariance-inducing weight sharing \citep{zhou2020meta}. However, these approaches do not aim to learn conserved quantities in dynamical systems.\looseness=-1

\vspace{-2mm}
\paragraph{Neural networks to discover physical laws.} There has been a growing interest in leveraging the functional search power of neural networks for better scientific models. Multiple works train a neural network and then analyze it to get information, such as inferring missing objects via back-propagation descent~\citep{alet2019neural}, inspecting its gradients to guide a program synthesis approach~\citep{udrescu2020ai}, learning a blue-print with a GNN~\citep{cranmer2020discovering}, or leveraging overparametrized-but-regularized auto-encoders~\citep{udrescu2021symbolic}. Others, such as DreamCoder~\citep{ellis2020dreamcoder}, take the explicit approach with a neural guided synthesis over the space of formulas. Unlike these works, our method can readily be applied both to symbolic conservation laws and to neural network conservation losses applied to raw videos.

\vspace{-1mm}
\section{Discussion}\label{sec:discussion}
\vspace{-1mm}
\paragraph{Limitations.}%\label{subsec:limitations}
A disadvantage of our proposed approach over architectural constraints (in the cases when the latter can represent the same bias) is computation time. Noether Networks optimize the conservation loss inside the prediction function, which results in an overhead. This can be reduced by only fine-tuning the higher-level layers, avoiding back-propagation back to the input. Another approach to reduce training time could be to warm-start the Noether embeddings with pre-trained contrastive embeddings, which have an aligned goal and have shown great results for video and audio classification~\citep{recasens2021broaden}. It is also sometimes the case that, by encoding the inductive bias during training the trained but un-tailored Noether Network performs better than a comparable architecture without the bias during training.
It is also worth noting that the theoretical guarantees apply to fully enforced conservation constraints, while in practice we only fine-tune our model to better satisfy them. However, we believe the theoretical result still conveys the potential benefits of Noether networks.

\paragraph{Broader impact.} %~\label{sec:impact}
We propose a new framework for meta-learning inductive biases in sequential settings, including from raw data. On the positive side, our framework can be used to discover conserved quantities in scientific applications, even when these are not fully conserved. By improving video prediction, we are also empowering predictive models for entities such as robots and self-driving cars, helping elder care and mobility. However, at the same time, it could be used for better predicting the movement and tracking of people from CCTV cameras, affecting privacy. It could also be used to improve the quality of deepfake video generation, which can be used for malicious purposes.

\paragraph{Overview and future work.} %~\label{subsec:overview}
We propose Noether Networks: a new framework for meta-learning inductive biases through conserved quantities imposed at prediction time. Originally motivated by applications in physics, our results indicate that Noether networks can recover known conservation laws in scientific data and also provide modest gains when applied to video prediction from raw pixel data. The generality of optimizing arbitrary unsupervised losses at prediction time provides an initial step towards broader applications.
Finally, this work points at the usefulness of designing meta-learned inductive biases by putting priors on the biases instead of hard-coding biases directly.

\section*{Acknowledgements}
We want to thank Leslie Pack Kaelbling, Tom\'as Lozano-P\'erez and Maria Bauza for insightful feedback.
We gratefully acknowledge support from GoodAI; from NSF grant 1723381; from AFOSR grant FA9550-17-1-0165; from ONR grants N00014-18-1-2847 and N00014-21-1-2685; from the Honda Research Institute, from MIT-IBM Watson Lab; from SUTD Temasek Laboratories; and from Google. Chelsea Finn is a fellow in the CIFAR Learning in Machines and Brains program. We also acknowledge the MIT SuperCloud and Lincoln Laboratory Supercomputing Center for providing HPC resources that have contributed to the reported research results. Any opinions, findings, and conclusions or recommendations expressed in this material are those of the authors and do not necessarily reflect the views of our sponsors.

\bibliography{main}

\newpage
\appendix

\section{Proofs} \label{app:proofs}
We utilize the following lemma in the proof of Theorem \ref{thm:2}. 
\begin{lemma} \label{lemma:1}
Fix $v \in \RR^d$.
Then, for any $\delta>0$,  with  probability at least $1-\delta$ over an
i.i.d.\ draw of $n$ examples $((x_i
, y_i))_{i=1}^n$, the following holds:
$$
\EE_{x,y}[\Lcal(f(x,v),y)]-\frac{1}{n} \sum_{i=1}^n\Lcal(f(x_i,v),y_i) \le C \sqrt{\frac{\ln (1/\delta)}{2n}}.  
$$ 
\end{lemma}
\begin{proof}[Proof of Lemma \ref{lemma:1}]
By using Hoeffding's inequality, we have that 
$$
\Pr\left(\EE_{x,y}[\Lcal(f(x),y)]-\frac{1}{n} \sum_{i=1}^n\Lcal(f(x_i,v),y_i) \ge  t \right) \le  \exp\left(-\frac{2nt^2}{C^2}\right),
$$
where $t \ge 0$. Solving $\delta=\exp\left(-\frac{2nt^2}{C^2}\right)$ for $t\ge 0$,
we have that 
for any $\delta>0$,  with  probability at least $1-\delta$, the following holds:
$$
\EE_{x,y}[\Lcal(f(x),y)]-\frac{1}{n} \sum_{i=1}^n\Lcal(f(x_i,v),y_i)\le C \sqrt{\frac{\ln (1/\delta)}{2n}}. 
$$ 

\end{proof}

\subsection{Proof of Theorem \ref{thm:2}}
\subsubsection{Preparation} \label{app:new:new:1}
In this subsection, we focus on the case of $g_\phi(f_{\theta}(x))\neq g_\phi(x)$ as a preparation for the more general case in the next subsection. The (closed) ball of radius $r$ centered at $c$ is denoted by $\Bcal_{r}[c]=\{v\in \RR^d : \|v- c \|_2\le r\}$. Fix $r > 0$ and  $\Ccal(r,\Vcal )\in \argmin_{\Ccal}\{|\Ccal|:\Ccal \subseteq\RR^d,\Vcal \subseteq \cup_{c \in\Ccal} \Bcal_{r}[c]\}$.
Let $\Ncal(r,\Vcal )=|\Ccal(r,\Vcal)|$; i.e. the minimum number of balls of radius $r$ needed to cover the a set of vectors $\Vcal$. 

The statement of theorem~\ref{thm:2} vacuously holds if $R$ is unbounded. Thus, we focus on the case of $R<\infty$ in the rest of the proof. 
For any $\theta \in \Theta$,
there exists $v\in \Vcal$ such that 
\begin{align}\label{eq:1}
\EE_{x,y}[\Lcal(f_{\theta}(x),y)]-\frac{1}{n} \sum_{i=1}^n\Lcal(f_\theta(x_i),y_i) =\EE_{x,y}[\Lcal(f(x,v),y)]-\frac{1}{n} \sum_{i=1}^n\Lcal(f(x_i,v),y_i)=\psi(v).
\end{align}
Moreover, for any  $v\in \Vcal$, the following holds: for any $c\in \Ccal(r,\Vcal )$,
\begin{align}\label{eq:2}
\psi(v)=\psi(c)+(\psi(v)-\psi(c)).
\end{align}
For the first term in the right-hand side of \eqref{eq:2}, by using Lemma \ref{lemma:1} with $\delta\rightarrow\delta/\Ncal(r,\Vcal )$ and taking union bounds, we have that for any $\delta>0$, with probability at least  $1-\delta$,
the following holds for all $c \in \Ccal(r,\Vcal )$: 
\begin{align}\label{eq:3}
\psi(c) \le C \sqrt{\frac{\ln (\Ncal(r,\Vcal )/\delta)}{2n}}. 
\end{align}
By combining equations \eqref{eq:2} and \eqref{eq:3}, we have that   for any $\delta>0$, with probability at least  $1-\delta$,
the following holds for any $v\in \Vcal$ and  all $c\in \Ccal(r,\Vcal )$:
\begin{align}
\psi(v)\le C \sqrt{\frac{\ln (\Ncal(r,\Vcal )/\delta)}{2n}}+(\psi(v)-\psi(c)).
\end{align}
This implies that for any $\delta>0$, with probability at least  $1-\delta$, the following holds for  any $v\in \Vcal$:
\begin{align} \label{eq:4}
\psi(v)\le C \sqrt{\frac{\ln (\Ncal(r,\Vcal )/\delta)}{2n}}+\min_{c \in \Ccal(r,\Vcal ) }|\psi(v)-\psi(c)|.
\end{align}
For the second term in the right-hand side of \eqref{eq:4}, we have that for  any $v\in \Vcal$,
\begin{align} \label{eq:4b}
\min_{c \in \Ccal(r,\Vcal ) }|\psi(v)-\psi(c)| \le \zeta\min_{c \in \Ccal(r,\Vcal ) } \|v-c\|_2 \le\zeta r. 
\end{align}
Thus, by using $r=\zeta^{1/\rho-1} \sqrt{\frac{1}{n}}$, we have that
for any $\delta>0$, with probability at least  $1-\delta$, the following holds for  all $v\in \Vcal$:
\begin{align} 
\psi(v)\le C \sqrt{\frac{\ln (\Ncal(r,\Vcal )/\delta)}{2n}}+ \sqrt{\frac{\zeta^{2/\rho}}{n}}.
\end{align}
Using equation \eqref{eq:1}, this implies that 
for any $\delta>0$, with probability at least  $1-\delta$, the following holds for  all $\theta \in \Theta$:
\begin{align} 
\EE_{x,y}[\Lcal(f_{\theta}(x),y)]-\frac{1}{n} \sum_{i=1}^n\Lcal(f_\theta(x_i),y_i)\le C \sqrt{\frac{\ln (\Ncal(r,\Vcal )/\delta)}{2n}}+ \sqrt{\frac{\zeta^{2/\rho}}{n}},
\end{align}
where $r=\zeta^{1/\rho-1} \sqrt{\frac{1}{n}}$. Since $\Ncal(r,\Vcal )\le (2R\sqrt d/r)^d= (2R(\zeta^{1-1/\rho})\sqrt{nd})^d$, 
\begin{align*} 
&\EE_{x,y}[\Lcal(f_{\theta}(x),y)]-\frac{1}{n} \sum_{i=1}^n\Lcal(f_\theta(x_i),y_i)
\\ &\le C \sqrt{\frac{\ln (\Ncal(r,\Vcal )/\delta)}{2n}}+ \sqrt{\frac{\zeta^{2/\rho}}{n}}
\\ & = C \sqrt{\frac{\ln (\Ncal(r,\Vcal ))+\ln(1/\delta)}{2n}}+ \sqrt{\frac{\zeta^{2/\rho}}{n}},
\\ & = C \sqrt{\frac{\ln ( (2R(\zeta^{1-1/\rho})\sqrt{nd})^d)+\ln(1/\delta)}{2n}}+ \sqrt{\frac{\zeta^{2/\rho}}{n}},
\\ & = C \sqrt{\frac{d\ln (\sqrt{d})+d\ln (2R(\zeta^{1-1/\rho})\sqrt{n})+\ln(1/\delta)}{2n}}+ \sqrt{\frac{\zeta^{2/\rho}}{n}}.
\end{align*}

\subsubsection{Putting results together}
In this subsection, we now generalize the proof of the previous subsection to the case of  $g_\phi(f_{\theta}(x))= g_\phi(x)$. The condition of $g_\phi(f_{\theta}(x))=g_\phi(x)$ implies that 
\begin{align}\label{eq:5}
g_\phi^{-1}[\{g_\phi(f_{\theta}(x))\}]=g_\phi^{-1}[\{g_\phi(x)\}].
\end{align}
Since $g_\phi^{-1}[\{g_\phi(x)\}]=\{x+Az : z \in \RR^m\}$ and $f_\theta(x)=x+v_\theta$, the right-hand side of equation \eqref{eq:5} can be simplified as
\begin{align*}
g_\phi^{-1}[\{g_\phi(f_{\theta}(x))\}]=\{f_{\theta}(x)+Az : z \in \RR^m\}=\{x+v_{\theta}+Az : z \in \RR^m\}.
\end{align*}
Substituting this into equation \eqref{eq:5} yields
\begin{align}\label{eq:6}
\{x+v_{\theta}+Az : z \in \RR^m\}=g_\phi^{-1}[\{g_\phi(x)\}]=\{x+Az : z \in \RR^m\},
\end{align}
where the last equality uses the fact that  $g_\phi^{-1}[\{g_\phi(x)\}]=\{x+Az : z \in \RR^m\}$. Equation \eqref{eq:6} implies that 
\begin{align}\label{eq:7}
v_{\theta} \in \Col (A) \subseteq \RR^d,
\end{align}
where $\Col (A)$ is the column space of the matrix $A \in \RR^{d\times m}$. Let  $\bA \in \RR^{d\times m}$ be a semi-orthogonal matrix
such that $\bA\T \bA=I_m$ (i.e., the identity matrix of size $m$ by $m$) and $ \Col (\bA) = \Col (A) $.  Then, equation \eqref{eq:7} implies that for any $\theta \in \Theta$, there exists $z \in \RR^m$ such that
\begin{align}\label{eq:8}
v_{\theta} = \bA z.
\end{align}We can further refine this statement by using the following observation. Since $\bA\T \bA=I_m$, we have that 
\begin{align}\label{eq:9}
R \ge \|v_\theta\|_2=\| \bA z\|_2= \|z\|_2,
\end{align}
and 
\begin{align} \label{eq:10}
|\psi(\bA z)-\psi(\bA z')|\le \ \zeta \|\bA (z-z')\|_2=\zeta \|z-z'\|_2.
\end{align}Define $\Zcal = \{z \in \RR^m: \|z\|_{2}\le R\}$. Then, equations \eqref{eq:7} and \eqref{eq:9} together imply that for any $\theta \in \Theta$, there exists $z \in \Zcal$ such that
\begin{align}\label{eq:11}
v_{\theta} = \bA z.
\end{align}
\iffalse
Whereas the previous subsection defines the ball in the space of $v \in \RR^d$, we can now define the ball in the space of $z \in \RR^m$ to cover the space of $v \in \RR^d$ through equation \eqref{eq:11}. That is, we define $\Bcal_{r}[c]=\{z\in \RR^m : \|z- c \|_2\le r\}$. Fix $r > 0$ and  $\Ccal(r,\Zcal )\in \argmin_{\Ccal}\{|\Ccal|:\Ccal \subseteq\RR^m,\Zcal\subseteq \cup_{c \in\Ccal} \Bcal_{r}[c]\}$. Let $\Ncal(r,\Zcal )=|\Ccal(r,\Zcal)|$. From here, the proof exactly follows the one from Section \ref{app:new:new:1} by replacing $v$ and $c \in \Ccal(r,\Vcal )$ with $\bA z$ and $\bA c$. Since the ball and the corresponding spaces are now defined in $\RR^m$ instead of $\RR^d$, this results in the bound with $m$ instead of $d$.

For the case $m=0$ we can get a tighter bound by noticing $\Ncal(r,\Zcal )=1$ with $r=0$ and using equations ~\ref{eq:4} and~\ref{eq:4b} in the proof in Section \ref{app:new:new:1}:
\begin{align*} 
\EE_{x,y}[\Lcal(f_{\theta}(x),y)]-\frac{1}{n} \sum_{i=1}^n\Lcal(f_\theta(x_i),y_i)
 \le C \sqrt{\frac{\ln (1/\delta)}{2n}}.
\end{align*} 
\fi

%\iffalse
Whereas the previous subsection defines the ball in the space of $v \in \RR^d$, we can now define the ball in the space of $z \in \RR^m$ to cover the space of $v \in \RR^d$ through equation \eqref{eq:11}. That is, we re-define  $\Bcal_{r}[c]=\{z\in \RR^m : \|z- c \|_2\le r\}$. Fix $r > 0$ and  $\Ccal(r,\Zcal )\in \argmin_{\Ccal}\{|\Ccal|:\Ccal \subseteq\RR^m,\Zcal\subseteq \cup_{c \in\Ccal} \Bcal_{r}[c]\}$.
Let $\Ncal(r,\Zcal )=|\Ccal(r,\Zcal)|$. Using 
\eqref{eq:11}, for any $\theta \in \Theta$,
there exists $z\in \Zcal$ such that 
\begin{align}\label{eq:12}
\nonumber \EE_{x,y}[\Lcal(f_{\theta}(x),y)]-\frac{1}{n} \sum_{i=1}^n\Lcal(f_\theta(x_i),y_i) &=\EE_{x,y}[\Lcal(f(x,\bA z),y)]-\frac{1}{n} \sum_{i=1}^n\Lcal(f(x_i,\bA z),y_i)
\\ & =\psi(\bA z),
\end{align}
Moreover, for any  $z\in \Zcal$, the following holds: for any $c\in \Ccal(r,\Zcal )$,
\begin{align}\label{eq:13}
\psi(\bA z)=\psi(\bA c)+(\psi(\bA z)-\psi(\bA c)).
\end{align}
For the first term in the right-hand side of \eqref{eq:13}, by using Lemma \ref{lemma:1} with $\delta\rightarrow\delta/\Ncal(r,\Zcal )$ and taking union bounds, we have that for any $\delta>0$, with probability at least  $1-\delta$,
the following holds for all $c \in \Ccal(r,\Zcal )$: 
\begin{align}\label{eq:14}
\psi(\bA c) \le C \sqrt{\frac{\ln (\Ncal(r,\Zcal )/\delta)}{2n}}. 
\end{align}
By combining equations \eqref{eq:13} and \eqref{eq:14}, we have that   for any $\delta>0$, with probability at least  $1-\delta$,
the following holds for any $z\in \Zcal$ and  all $c\in \Ccal(r,\Zcal )$:
\begin{align}
\psi(\bA z)\le C \sqrt{\frac{\ln (\Ncal(r,\Zcal )/\delta)}{2n}}+(\psi(\bA z)-\psi(\bA c)).
\end{align}
This implies that for any $\delta>0$, with probability at least  $1-\delta$, the following holds for  any $z\in \Zcal$:
\begin{align} \label{eq:15}
\psi(\bA z)\le C \sqrt{\frac{\ln (\Ncal(r,\Zcal )/\delta)}{2n}}+\min_{c \in \Ccal(r,\Zcal ) }|\psi(\bA z)-\psi(\bA c)|.
\end{align}
For the second term in the right-hand side of \eqref{eq:15}, by using equation \eqref{eq:10}, we have that for  any $z\in \Zcal$,
$$
\min_{c \in \Ccal(r,\Zcal ) }|\psi(\bA z)-\psi(\bA c)| \le \zeta\min_{c \in \Ccal(r,\Vcal ) } \|z-c\|_2 \le\zeta r. 
$$
If $m=0$, then since $\Ncal(r,\Zcal )=1$ with $r=0$, 
\begin{align*} 
\EE_{x,y}[\Lcal(f_{\theta}(x),y)]-\frac{1}{n} \sum_{i=1}^n\Lcal(f_\theta(x_i),y_i)
 \le C \sqrt{\frac{\ln (1/\delta)}{2n}}.
\end{align*}
This proves the desired statement for $m=0$. Thus, we focus on the case of $m \ge 1$ in the rest of the proof.
By using $r=\zeta^{1/\rho-1} \sqrt{\frac{1}{n}}$, we have that
for any $\delta>0$, with probability at least  $1-\delta$, the following holds for  all $z\in \Zcal$:
\begin{align} 
\psi(\bA z)\le C \sqrt{\frac{\ln (\Ncal(r,\Zcal )/\delta)}{2n}}+ \sqrt{\frac{\zeta^{2/\rho}}{n}}.
\end{align}
Using equation \eqref{eq:12}, this implies that 
for any $\delta>0$, with probability at least  $1-\delta$, the following holds for  all $\theta \in \Theta$:
\begin{align} 
\EE_{x,y}[\Lcal(f_{\theta}(x),y)]-\frac{1}{n} \sum_{i=1}^n\Lcal(f_\theta(x_i),y_i)\le C \sqrt{\frac{\ln (\Ncal(r,\Zcal )/\delta)}{2n}}+ \sqrt{\frac{\zeta^{2/\rho}}{n}},
\end{align}
where $r=\zeta^{1/\rho-1} \sqrt{\frac{1}{n}}$. Since $\Ncal(r,\Zcal )\le (2R\sqrt m/r)^m= (2R(\zeta^{1-1/\rho})\sqrt{n m})^m$, 
\begin{align*} 
&\EE_{x,y}[\Lcal(f_{\theta}(x),y)]-\frac{1}{n} \sum_{i=1}^n\Lcal(f_\theta(x_i),y_i)
\\ &\le C \sqrt{\frac{\ln (\Ncal(r,\Zcal )/\delta)}{2n}}+ \sqrt{\frac{\zeta^{2/\rho}}{n}}
\\ & = C \sqrt{\frac{\ln (\Ncal(r,\Zcal ))+\ln(1/\delta)}{2n}}+ \sqrt{\frac{\zeta^{2/\rho}}{n}},
\\ & = C \sqrt{\frac{\ln ( (2R(\zeta^{1-1/\rho})\sqrt{nm})^m)+\ln(1/\delta)}{2n}}+ \sqrt{\frac{\zeta^{2/\rho}}{n}},
\\ & = C \sqrt{\frac{m\ln (\sqrt{m})+m\ln (2R(\zeta^{1-1/\rho})\sqrt{n})+\ln(1/\delta)}{2n}}+ \sqrt{\frac{\zeta^{2/\rho}}{n}}.
\end{align*}
%\fi
\qed

\section{Why Noether Networks avoid learning trivial conserved quantities}~\label{app:trivial}
Neural networks have large capacity. Therefore, we have to be careful that the Noether embedding is not learning a quantity that is trivially conserved but is unrelated to the task, such as predicting a constant vector $g_\phi(x)=\vec{C}$. In our setup both the prediction network and the meta-learned conservation embedding are trained end-to-end to minimize the task loss after the conservation update. However, for the meta-learned embedding this is equivalent to encouraging a big decrease in supervised loss after the update than before the update. This can be seen by looking at the gradient of the difference between the task loss before and after the update:
$$\nabla_\phi\left(\mathcal{L}_{\rm task}(x_{1:T},f_{\theta(x_0;\phi)})-\mathcal{L}_{\rm task}(x_{1:T},f_{\theta})\right)$$
Here $\theta(x_0;\phi)$ denotes the model parameters after the inner loss update which conserves the embedding $g_\phi$ for input $x_0$. Notice that the second loss does not depend on the embedding (since $\theta$ has not yet been tailored with $g_\phi$). Thus, the gradient of the overall expression is equivalent to:
$$\nabla_\phi\left(\mathcal{L}_{\rm task}(x_{1:T},f_{\theta(x_0;\phi)})\right)$$
which is the one we optimize. A trivially conserved quantity (like $g_\phi(x)=0$ independent of $\phi$ and $x$) will not produce any improvement on the supervised loss after its conservation is encouraged. Therefore, the embedding will seek to be more useful, by making the tailored weights $\theta(x_0;\phi)$ perform better than the untailored ones $\theta$.
\section{Experimental details for scientific data}\label{app:scientific}

We use three different experiments from~\citep{greydanus2019hamiltonian}: the ideal pendulum, the ideal spring, and the real pendulum, under an Apache license. The first two come from simulated ODEs without noise, the latter comes from a real data from~\citet{schmidt2009distilling}. It is worth noting that other datasets from~\citet{greydanus2019hamiltonian} (such as a planetary system) could not be included because the DSL required too much depth to reach the conserved energy quantity. A better search, such as using evolution, or a better DSL (such as those derived from many scientific formulas in DreamCoder~\citep{ellis2020dreamcoder}) could remedy this. Finally, note that an approach that searched over the same DSL encoding the loss as a generic $f(x_0,x_t)$ loss, not as a conservation $|g(x_0)-g(x_t)|$ would require more than twice the depth, thus not being able to cover the evaluated datasets.

Most experimental details and hyperparameters are already detailed in the main text. It is worth noting that the pipeline approach: first discovering approximately conserved quantities, then pruning a subset of 20 that are most useful  for tailoring and then finally pick the best loss when used with meta-tailoring. This pipeline, along with the concrete numbers of 20 candidates and 2 orders of magnitude comparing variance in real data vs. random data, were heuristically chosen to speed up the search, without trying other hyper-parameters. For the real pendulum we found that two losses that were among the top 20 for tailoring diverged when used for meta-tailoring. We conjecture this is the case because of the noise in the real data.

Experiments were performed on an NVIDIA Tesla V-100 GPU and 10 CPU cores, taking around 4 hours to run.

\paragraph{Extra details on state space and energies} For the pendulum, input $x=(p,q)\in\mathbb{R}^2$ contains its angle $q$ and momentum $p$. The formula for the energy is $\mathcal{H} = 2mgl(1-\cos{q}) + \frac{l^2p^2}{2m}$. \citet{greydanus2019hamiltonian} set $m=\frac{1}{2},l=1,g=3$, resulting in $\mathcal{H} = 3(1-\cos{q}) + p^2$. Notice that a simpler conserved quantity is $p^2 - 3\cdot\cos{q}$.
For the spring, input $x=(p,q)\in\mathbb{R}^2$ contains the displacement $q$ and momentum $p$, and the system's energy is given by $\mathcal{H} = \frac{1}{2}kq^2 + \frac{p^2}{2m}$. \citet{greydanus2019hamiltonian} set $k=q=m=1$, resulting in $\mathcal{H} = \frac{1}{2}(q^2 + p^2)$, where units are omitted. Thus, $q^2+1\cdot p^2$ is a conserved quantity, where coefficient 1 has appropriate units.

For the real pendulum, it is worth noting that the energy candidate $p^2-2.4\cos{q}$ was proposed by humans to fit the data, since the data is experimental. It could thus be slightly sub-optimal, which could explain why Noether Networks improves its predictions. We also consider the possibility of it being due to fluctuations in noisy data and it being a small dataset.

\section{Experimental Details for pendulum with controls}
We generated the action-conditioned video prediction data in OpenAI Gym's ``Pendulum-v0'' environment. We want to make each episode different and not easily predictable, but also not too erratic. Hence we generate actions for each episode by:
\begin{equation*}
a_t = 2 \sin(\omega t + \alpha)
\end{equation*}
where $\omega,\alpha$ are randomly sampled for each episode, with $\omega\sim \text{U}[0.05, 0.15]$ and $\alpha\sim\text{U}[0, 2\pi]$. We recorded $200$ episodes of $200$ timesteps each and split them into $5$ train and $195$ test videos.

Our action conditioned video prediction architecture is based off of the publicly available implementation\footnote{\href{https://github.com/edenton/svg}{https://github.com/edenton/svg}} of SVG, with prior sampling removed and actions concatenated to the output of the frame encoder. The frame encoder and decoder are based on the DCGAN generator and discriminator, and the latent frame predictor is a 2-layer LSTM with hidden size 256. We resize the input frames to $64\times 64$ before inputting to the frame encoder. Throughout all experiments we use a batch size of $32$ and optimize using Adam with a learning rate of $0.001$.

During evaluation, each method receives a length-$4$ frame history and $26$ future actions and must predict the next $26$ frames. However, during training each method only predicts the next $10$ frames and then is updated to minimize mean square error; this speeds up each step and reduces memory cost. For the Noether Network there is an inner loop where we minimize the learned conservation loss: in these experiments, we found it better to update the LSTM's initial outputs during the inner loop, rather than any network parameters. We then pass the updated LSTM outputs into the frame decoder to generate a video prediction. The inner loop optimization is done with gradient descent using a learning rate of $0.5$, for $1$ step. As described above the conservation loss involves conserving a learned embedding over time. Our learned embedding in this case is a $3$-layer MLP with $512$ units at each hidden layer and outputs an embedding with dimension $100$. The embedding receives as input the LSTM output for the past two time steps and the most recent action, all concatenated together.

We ran all the action conditioned experiments on a single machine with an NVIDIA RTX 2080 Ti GPU. Due to the limited data available, all training was relatively fast: the initial 50 epochs of SVG training took $\sim 40$ minutes, while the Noether Network tuning process took $\sim 17$ minutes.

\section{Experimental Details for Physics 101}
We use a subset of the Physics 101 dataset \citep{wu2016physics} with videos from the ramp scenario, where various objects are let go by a human hand at the top of a ramp. We could not find its associated license. There are two ramp settings: 10 degrees and 20 degrees.  To ensure the prediction problem is interesting and nontrivial, we only take examples with the 20 degree ramp since the object does not slide down the ramp in many of the 10 degree examples.  Additionally, we only take sequences with length at least 2 seconds, and use a frame rate of 15 frames per second when extracting frames to pass to the models.  We use videos recorded with the \verb|Kinect_RGB_1| sensor.  Video frames are center-cropped at full height (1080 pixels) and downsample to obtain $128\times 128$ images, and perform random horizontal flipping for sequences.  This preprocessing results in 389 possible sequences, and we take a random 80/20 train/test split.  In the prediction task, we condition the model on two frames, and the model must predict the subsequent 20 frames.  To ensure that we extract segments of videos where the object is moving, we take the middle 22 frames of each video clip.  We use a mini-batch size of 2 for these experiments.

Our baseline model is based on the publicly-available implementation of SVG-LP \cite{denton2018stochastic}.  As in the original implementation, the encoder and decoder are based on VGG16 \cite{simonyan2014very}, the latent frame predictor is a 2-layer LSTM with hidden size of 128, and both the prior and posterior are single-layer LSTMs that produce latent variables $\mathbf{z}_t$ of dimension 64.  We replace all batch normalization layers with layer normalization layers in both the baseline and the Noether Network because of the small mini-batch size used in our experiments. Preliminary experiments showed similar performance, but much more stable training. The learning rate used during the training of the baseline $0.0001$.  Teacher forcing is used during the training of the SVG baseline and the meta-training of the Noether Network, as done by \citet{denton2018stochastic}.

The meta-learned inner loss of the Noether Network is formulated as in equation~\ref{eq:noether_zero_loss}(b), with the learned embedding $g_\phi$ parameterized as a 2-layer convolutional network where each layer consists of a $5\times 5$ convolution with 32 filters in the first layer and 64 filters in the second layer, a ReLU non-linearity, and $2\times 2$ max pooling.  There is a final linear layer which projects onto a 64-dimensional embedding space.

The \textsc{CNGrad} algorithm of \citet{alet2020tailoring} is used to perform tailoring.  In the Noether Network, conditional normalization (CN) layers are inserted after each batch normalization layer in both the encoder and the decoder.  The CN parameters are initialized to the identity transformation, and are adapted with a single inner step at prediction-time using an inner learning rate of $0.0001$.  The embedding is trained in the outer loop with an outer learning rate of $0.0001$.  We report results obtained with meta-tailoring, where both the base model and the embedding are randomly initialized.

We tried multiple Noether embedding dimensions as well as inner learning rates, but results were very consistent across these choices so we left our original choices. The only sets which failed to learn were very low inner learning rates ($\leq 10^{-5}$) with very high ($>1024$) embedding dimensions. We conjecture this stability is due to two reasons:
\begin{enumerate}
    \item Meta-learning the loss function compensates for the choice of inner learning rate: for instance we observed that increasing the inner learning rate by two orders of magnitude resulted in two orders of magnitude smaller losses (and thus equal gradient updates).
    \item PCA analysis of the Noether embeddings suggests they fundamentally encoded few ($\approx 4$) dimensions, even when the embeddings had more dimensions. More work needs to be done to understand whether this is a fundamental property of the domain discovered by the method or an optimization issue.
\end{enumerate}

We ran all experiments with the Physics 101 dataset on an NVIDIA Tesla V100 GPU and 20 CPU cores.  Training the baseline until convergence took 400 epochs and approximately 13 hours, and training the Noether Network took approximately 2 days, 23 hours. The error bars reported in figure~\ref{fig:phys101-metrics} and figure~\ref{fig:phys101-inner-outer} are standard error of the mean (SEM) for $N=5$ training seeds.

\section{Grad-CAM localization maps}\label{app:grad_cam}
\begin{figure}[h]
    \centering
    \includegraphics[width=\textwidth]{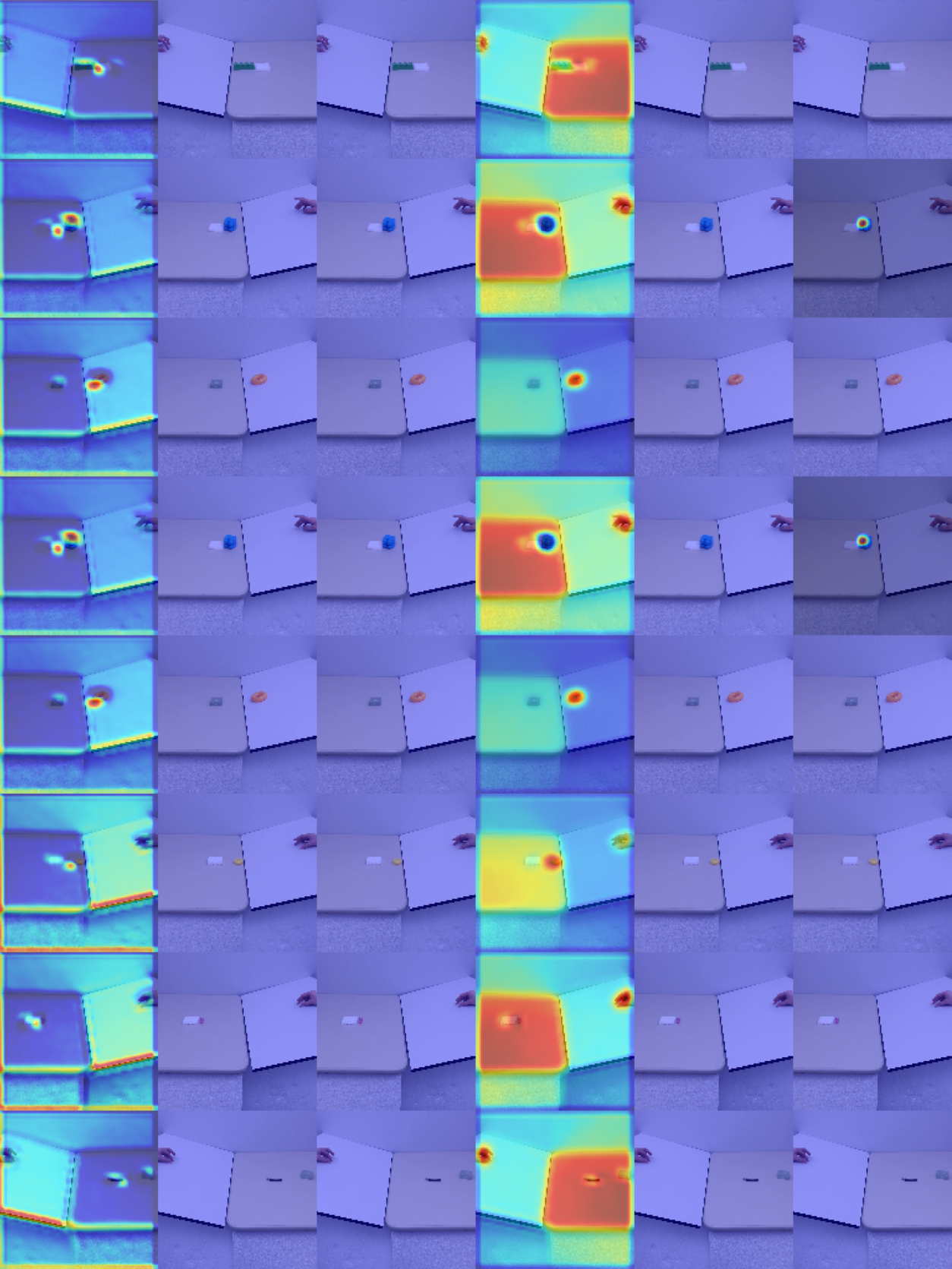}
    Embedding dimension (after PCA) $\longrightarrow$
    \caption{Grad-CAM localization maps have consistent behavior across sequences in the test set. Each row corresponds to a different sequence, and each column represents an embedding dimension (first PCA dimension on the left).}
    \label{fig:phys101-grad_cam_more_examples}
\end{figure}

In figure~\ref{fig:phys101-grad_cam_more_examples}, Grad-CAM heatmaps are shown for a single time step from each of five example sequences from the test set. The trends discussed in section~\ref{sec:evaluation} are consistent across these examples as well.

\end{document}